\documentclass{article}

\usepackage{graphicx}
\usepackage{amsmath}
\usepackage{amssymb}
\usepackage{algorithm}
\usepackage{algpseudocode}
\usepackage{float}
\usepackage{wrapfig}
\usepackage{booktabs}
\usepackage{multirow}
\usepackage{float}
\usepackage{etoc}
\usepackage{subcaption}

\usepackage[main, final]{neurips_2026}

\usepackage[utf8]{inputenc} %
\usepackage[T1]{fontenc}    %
\usepackage{hyperref}       %
\usepackage{url}            %
\usepackage{booktabs}       %
\usepackage{amsfonts}       %
\usepackage{nicefrac}       %
\usepackage{microtype}      %
\usepackage{xcolor}         %

\title{Flow-Corrected Shape Optimization: Taming Manifold Drift in High-Dimensional 3D Models}

\author{%
  Emilien Seiler \\
  CVLab, EPFL \\
  \texttt{emilien.seiler@epfl.ch} \\
  \And
  Nicolas Talabot \\
  CVLab, EPFL \\
  \texttt{nicolas.talabot@epfl.ch} \\
  \And
  Yingxuan You \\
  CVLab, EPFL \\
  \texttt{yingxuan.you@epfl.ch} \\
  \And
  Federico Stella \\
  CVLab, EPFL \\
  \texttt{federico.stella@epfl.ch} \\
  \And
  Pascal Fua \\
  CVLab, EPFL \\
  \texttt{pascal.fua@epfl.ch} \\
}

\newif\ifdraft
\draftfalse

\ifdraft
\newcommand{\PF}[1]{{\color{red}{\bf PF: #1}}}

\newcommand{\ES}[1]{{\color{blue}{\bf ES: #1}}}

\newcommand{\NT}[1]{{\color{violet}{\bf NT: #1}}}

\newcommand{\FS}[1]{{\color{orange}{\bf FS: #1}}}

\newcommand{\YY}[1]{{\color{brown}{\bf YY: #1}}}

\else
\newcommand{\PF}[1]{}

\newcommand{\ES}[1]{}

\newcommand{\NT}[1]{}

\newcommand{\FS}[1]{}

\newcommand{\YY}[1]{}

\fi

\newcommand{\parag}[1]{\vspace{-3mm}\paragraph{#1}}

\newcommand{\bz}{\mathbf{z}}

\newcommand{\real}{\mathbb{R}}

\DeclareMathOperator*{\argmin}{arg\,min}

\newcommand{\cM}{\mathcal{M}}
\newcommand{\cO}{\mathcal{O}}

\newcommand{\ours}{FCSO}

\newlength{\mytabcolsep}
\makeatletter
\DeclareRobustCommand\onedot{\futurelet\@let@token\@onedot}
\def\@onedot{\ifx\@let@token.\else.\null\fi\xspace}

\makeatother

\begin{document}

\maketitle


\begin{abstract}

Optimizing 3D shapes within the latent spaces of deep generative models is fundamental to computer assisted engineering, yet remains prone to a critical failure mode we term \textit{manifold drift}: the tendency of gradient-based optimization to move latent vectors away from the manifold of valid shapes.
This problem is exacerbated in state-of-the-art 3D shape generative models that operate in increasingly high-dimensional latent spaces where valid shapes occupy a vanishingly small fraction of the full space.
Existing mitigation strategies, including latent regularization and flow-matching approaches, either sacrifice expressiveness, demand a difficult trade-off between objective guidance and generative fidelity that remains prone to manifold drift, or are computationally infeasible to scale to modern, large-capacity 3D shape models.
We introduce a novel optimizer-corrector framework that alternates between gradient steps for objective minimization and guided flow matching to drive the latent state back to the valid shape manifold.
By decoupling objective minimization from flow-based correction, optimizing freely and correcting strictly, this alternating design avoids inherent trade-offs, preserving geometric validity without sacrificing expressiveness while remaining computationally feasible on modern 3D shape models.
We demonstrate its effectiveness across generative priors of varying complexity, from simple vector latent spaces to large-scale architectures across a variety of downstream optimization tasks, including aerodynamic drag reduction and object compliance optimization.

\end{abstract}


\section{Introduction}
\label{sec:intro}

Refining 3D shapes to maximize a measure of their performance is central to many engineering disciplines, from Computer Assisted Design to Computer Vision and Graphics, with applications in aeronautics, healthcare, and manufacturing~\cite{Lopez24a,Zhu21a,Iqbal19a,Tan21a}.
With the advent of generative models, the approaches have shifted from traditional mesh-based refinement to continuous optimization within learned latent spaces~\cite{Wei24a,Chen25f,Chen26a}.
To capture finer geometric details, state-of-the-art shape models~\cite{Yang25a, Zhang23d, Xiang25a} have grown large and now operate in very high-dimensional latent spaces.

\begin{figure}[t]
    \centering
    \captionsetup[figure]{skip=0.1pt}
    \begin{subfigure}[b]{0.48\textwidth}
        \textbf{(a)}~\raisebox{-0.5\height}{\includegraphics[height=4cm]{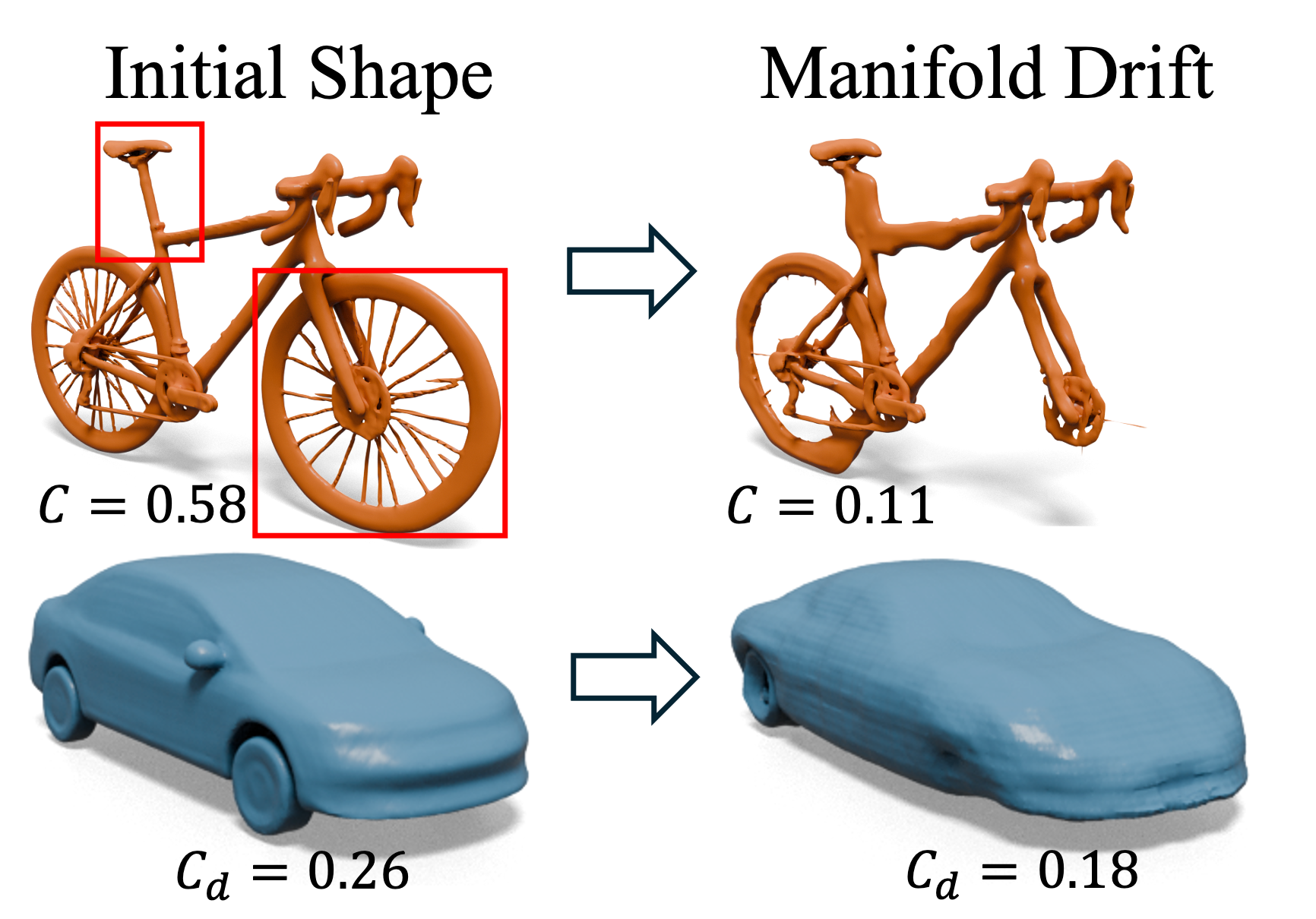}}
        \label{fig:teaser_a}
    \end{subfigure}
    \hfill 
    \begin{subfigure}[b]{0.48\textwidth}
        \textbf{(b)}~\raisebox{-0.5\height}{\includegraphics[height=4cm]{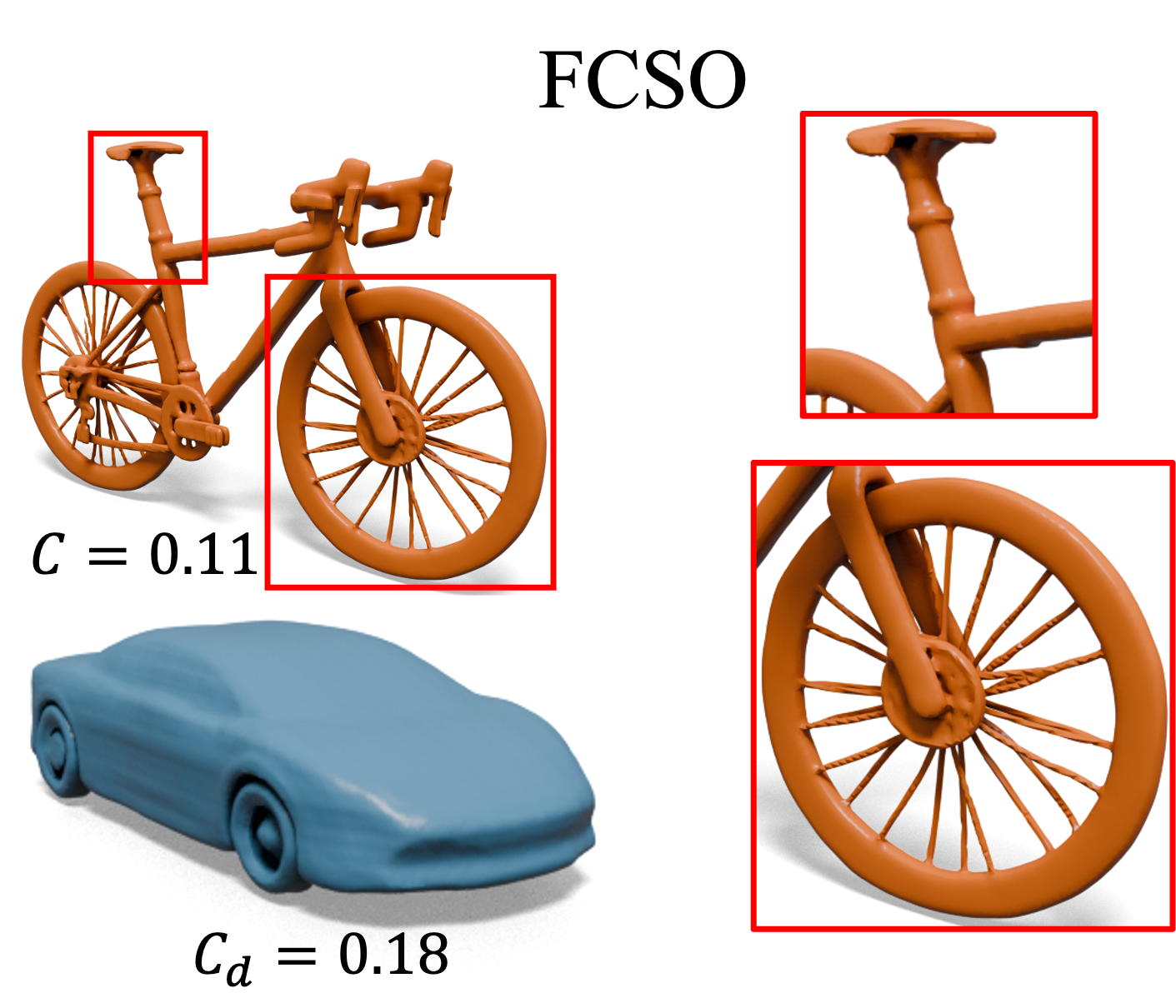}}
        \label{fig:teaser_b}
    \end{subfigure}

    \vspace{0.0cm} %

    \begin{subfigure}[b]{0.98\textwidth}
        \textbf{(c)}~\raisebox{-0.5\height}{\includegraphics[width=0.92\textwidth]{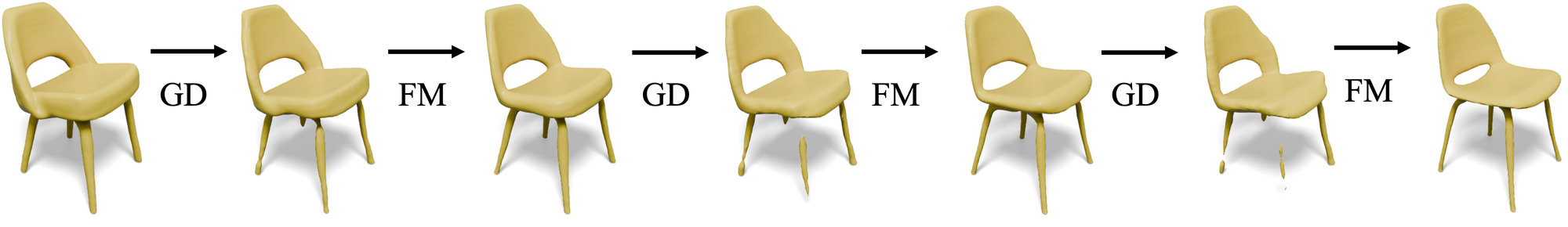}}
        \label{fig:teaser_c}
    \end{subfigure}
    
    \caption{Visualizing (a) manifold drift in gradient descent optimization for bike stiffness and car drag vs. (b) our results on the same tasks. (c) Our proposed framework, which alternates between Gradient Descent (GD) and Flow Matching (FM) correction, applied here to volume reduction.}
    \label{fig:teaser_main}
    \vspace{-5mm}
\end{figure}

While this increase in complexity has boosted both descriptive power and accuracy for shape {\it reconstruction} and \textit{generation}, it also introduces new challenges for shape {\it optimization}.
In practice, one often starts with a shape that is within a training distribution. This distribution often lives on a manifold occupying only a small part of the complete latent space.
The initial shape is then refined to minimize an objective function, with no guarantee that it will stay close to that manifold.
Thus, as the optimization progresses, the shape may end up far from it, making it potentially nonsensical as shown in Fig.~\ref{fig:teaser_main}(a), a phenomenon we will refer to as {\it manifold drift}.
This problem was already observed when using early neural representations~\cite{Remelli20b} but is more acute for new, larger ones due to the {\it Curse of Dimensionality}.
In high-dimensional spaces, the manifold of valid shapes occupies an ever smaller fraction of the complete space, making manifold drift even more of an issue. 


As this is a critical vulnerability of generative methods, mitigation strategies have been proposed.
Latent regularization techniques typically trade off the expressiveness of the search space for geometric validity~\cite{Liu22g,Samuel23a}.
More recent ones rely on flow matching~\cite{Lipman22} and diffusion models~\cite{Song21c}.
Some approaches optimize directly with respect to the starting point of the flow matching process, which is effective for small models but quickly becomes computationally intractable when using modern 3D architectures~\cite{BenHamu24, Tang24b, Chen25f}.
Other approaches use the optimization objective to provide guidance at inference time~\cite{Guo24f,Zhang25e}. 
However, because these methods attempt to perform both objective optimization and geometric correction in a single pass, they force a complex trade-off.
Too much guidance breaks the generative prior, causing manifold drift and degenerate shapes, while too little guidance results in failure to meaningfully optimize the objective.

In this paper, we address these issues by introducing Flow-Corrected Shape Optimization (\ours{}), a novel framework that explicitly separates the gradient descent from the correction steps to ensure robust convergence while remaining computationally tractable, effectively preventing manifold drift, as shown in Fig.~\ref{fig:teaser_main}(b).
Conceptually, our algorithm operates as an optimizer-corrector mechanism.
The gradient descent steps act as the optimizer that moves the latent vector strictly in the direction of steepest descent to minimize the objective.
The flow matching acts as the corrector, driving the updated state back to the valid shape manifold.
As illustrated in Fig.~\ref{fig:teaser_main}(c), by alternating these steps, we make progress on the optimization landscape without accumulating geometric errors over successive iterations.
We can leverage pre-trained flow models when provided, without requiring any retraining.

We demonstrate the effectiveness and scalability of \ours{} with generative shape priors, from simple vector latent spaces~\cite{Park19c} to newer much more complex ones~\cite{Yang25a}. Thus, our contribution is twofold:
\begin{itemize}
	\vspace{-2mm}
	 \item We show that the increased expressivity of modern 3D models fundamentally degrades the reliability of gradient-based latent optimization and that current diffusion and flow-based models are insufficient to mitigate this.
	 \item We introduce a robust shape optimization algorithm that alternates between gradient-based objective minimization and flow-matching corrections.
       This approach effectively addresses the manifold drift problem while remaining computationally tractable, even when scaling to highly complex architectures.
\end{itemize}
\vspace{-2mm}
Our experimental results show that we outperform state-of-the-art baselines on several representative shape optimization tasks; the code and benchmark will be released at \url{https://github.com/eseiler18/fcso}.


\section{Related work}
\label{sec:related}

\subsection{Shape modeling}
Recent advances in 3D generative models leverage expressive latent spaces to map compact latent vectors into complex 3D geometries~\cite{Dong24a, Zeng22, Gao22a}.
These vectors can be decoded into various representations, such as voxel grids~\cite{Wu15b}, point clouds~\cite{Fan17a}, or implicit neural fields~\cite{Park19c,Mescheder19}.
The latter continuously describe surfaces that can subsequently be extracted as explicit meshes~\cite{Lorensen87,Shen21a,Stella25}.
To deliver higher-fidelity and more precise 3D representation, the complexity of these latent spaces has significantly increased.
Early approaches, such as DeepSDF~\cite{Park19c}, relied on a single global latent vector to represent an entire shape and, while effective for simple topologies, this makes it difficult to capture fine details.
More recent models, such as 3DShape2VecSet~\cite{Zhang23d}, Dora~\cite{Chen25b} or Hunyuan3D~\cite{Yang25a}, replace the single vectors by sets of vectors and incorporate attention layers into the decoders to increase expressivity.
On the other hand, local latent representations in the form of sparse featurized voxels allow each vector to locally represent a specific region of the shape~\cite{Xiang25a, He25a}.
Driven by scaling laws, these recent architectures require a high-dimensional parameter space and massive training datasets.
For example, Hunyuan3D~\cite{Yang25a} represents a single shape using 4,096 vectors of size 64, offering state-of-the-art geometric fidelity.
While complex shape priors are essential for high-quality representations, they exacerbate manifold drift in optimization as the set of latents that decode into meaningful shapes becomes increasingly sparse.
We propose a solution that mitigates this for arbitrary shape priors of any complexity as long as a differentiable mapping exists between the latent representation and the optimization objective.

\subsection{Gradient-based optimization}
\label{sec:optim}
Because the sophisticated models discussed above have so many parameters, black-box optimization, including Bayesian optimization~\cite{Shahriari16a} and evolutionary strategies~\cite{Hansen16a}, are not always practical and gradient-based methods are better suited.
Thus, the gradients have to be computed and back-propagated to the latent parameters~\cite{Zhan25a, Remelli20b}.
While some objectives can be computed directly from the shape or the neural field~\cite{Wei24a, Zhan25a}, others depend on complex physical quantities estimated using numerical simulations that are both computationally expensive and non-differentiable~\cite{OpenFoam}.
A common workaround is then to train surrogate models based on Gaussian Processes or Neural Networks~\cite{Baque18,Guillard24a,Wu24a} to predict these physical quantities.
The former are well established in many engineering disciplines under the name of {\it Kriging}~\cite{Jones98a} but are less than ideal to deal with large numbers of parameters.
The latter can handle high-dimensional latent vectors, enabling fast gradient evaluation via back-propagation through the surrogate network.
Another approach is to integrate physical surrogates directly as auxiliary decoders operating on the latent space, coupling shape generation and physical prediction~\cite{You26a}.
However, as observed in recent studies~\cite{Guillard24a,Chen25f}, optimizing latent vectors inherently pushes them away from the manifold of valid shapes, yielding shapes that are invalid even if they minimize the objective.
This is compounded by the use of surrogate objective functions, which become unreliable out-of-distribution and can erroneously assign high scores to these degenerate shapes.
This challenge echoes the out-of-domain artifacts identified in 2D GAN inversion~\cite{Zhu20, Tov21}, those early 2D works used adversarial regularizers to keep latent codes valid, they inherit standard GAN instabilities.

\subsection{Flow matching for optimization}
Diffusion~\cite{Song21c} and Flow Matching~\cite{Lipman22} have emerged as promising ways to provide the necessary control by bringing back the drifting data to the training distribution.
Related work in image restoration often uses diffusion or flow models for inverse problems~\cite{BenHamu24, Chung24a, Kim25a, Zhang25e, Patel24a}. Like us, they minimize an objective while maintaining meaningful outputs.
However, these methods typically rely on a data-driven reconstruction term to anchor the generation process~\cite{Kawar22a}, mitigating manifold drift, however, no such reference exists in our setting.
One approach differentiates directly through the flow process by back-propagating the objective from the final output to the initial noise vector~\cite{BenHamu24, Chen25f, Tang24b}.
While mathematically elegant, back-propagating through an entire ODE solver incurs massive overheads, scaling poorly to high-dimensional latent spaces. Moreover, our experiments show that even these intensive approaches fail to prevent manifold drift.
To avoid the heavy back-propagation cost, one can instead rely on gradient guidance within the generation process.
Recent work has explored optimization via guided diffusion~\cite{Guo24f} extended to flow matching models by incorporating a trajectory regularization guided by the score function and the trace of the velocity field’s Jacobian~\cite{Zhang25e}.
However, guidance methods require sensitive hyper-parameter tuning to balance a fundamental trade-off: while strong guidance causes manifold drift by breaking the generative prior, weak guidance fails to meaningfully minimize the objective.
Yet another approach is to introduce a stochastic score approximation regularization during the optimization~\cite{Chen26a} but relying on soft regularization often fails to strictly prevent manifold drift in highly complex spaces.
In contrast, alternating between gradient and flow matching steps, as we propose, effectively handles manifold drift without any of the above drawbacks.


\section{Background: flow matching}
\label{sec:preliminaries}

We briefly review the standard Flow Matching (FM)~\cite{Lipman22,Albergo23a,Liu23e} approach.

Following the convention, Gaussian noise is sampled at time $t=0$ and de-noised. The clean data is obtained at time $t=1$.
To this end, FM constructs a probability density path $p_t(\mathbf{x})$ that transforms a simple prior $p_0$ ($\mathcal{N}(\mathbf{0}, \mathbf{I})$ in the case of Gaussian noise) into a complex data distribution $p_1$ over time $t \in [0, 1]$.
This transformation is governed by an Ordinary Differential Equation (ODE)

\begin{equation}
    \frac{d\mathbf{x}_t}{dt} = \mathbf{v}(\mathbf{x}_t), \quad \mbox{with} \quad \mathbf{x}_0 \sim p_0,
\end{equation}

where $\mathbf{v}$ is the vector field.
A neural network $\mathbf{v}_\theta$, typically a U-Net~\cite{Ronneberger15} or a Transformer~\cite{Esser24a,Vaswani17}, is trained to estimate $\mathbf{v}$ from $\mathbf{x}_t$ and $t$.
As doing so for all possible values of $\mathbf{x}$ would be intractable, we use the Rectified Flow formulation~\cite{Liu23e}, which relies on the straight-line interpolation between sampled noise $\mathbf{x}_0$ and data point $\mathbf{x}_1$

\begin{equation}
    \mathbf{x}_t = (1-t)\mathbf{x}_0 + t\mathbf{x}_1 \; .
\end{equation}

The target conditional vector field reduces to the constant velocity $\mathbf{v}(\mathbf{x}_t | \mathbf{x}_0, \mathbf{x}_1) = \mathbf{x}_1 - \mathbf{x}_0$ and the network is trained by minimizing the loss

\begin{equation}
    \mathcal{L}_{\text{FM}}(\theta) = \mathbb{E}_{t, \mathbf{x}_0, \mathbf{x}_1} \left[ \left\| \mathbf{v}_\theta(\mathbf{x}_t, t) - (\mathbf{x}_1 - \mathbf{x}_0) \right\|^2 \right].
\end{equation}


\section{Method}
\label{sec:method}

Our goal is to refine a 3D shape $S$ to satisfy a specific numerical objective, while leveraging known geometric priors learned by a pre-trained generative model.
We assume that we are given a pre-trained decoder $\mathcal{D}: \mathcal{Z} \rightarrow \mathcal{S}$ that maps a latent $\bz$ to its explicit 3D spatial representation $S = \mathcal{D}(\bz)$.
Given  a task-specific objective function $\cO: \mathcal{S}\to\real$ and an initial latent vector $\bz_s$, our task is to minimize $\cO$ with respect to $\bz$ to find
\begin{equation}
    \label{eq:opt_obj}
    \mathbf{z}_s^* = \argmin_{\mathbf{z}} \mathcal{O}\left(\mathcal{D}(\mathbf{z})\right), \quad \text{initialized at } \mathbf{z}^{(0)} = \mathbf{z}_s \; .
\end{equation}
In practice, $\bz$ can be either a single global vector as in DeepSDF~\cite{Park19c}, or a set of latent vectors as in 3DShape2VecSet~\cite{Zhang23d} or Hunyuan3D~\cite{Yang25a}.

\subsection{The manifold drift challenge}
\label{sec:manifold_drift}

While seemingly straightforward in theory, the optimization of Eq.~\ref{eq:opt_obj} is fraught with problems in practice:
As discussed in Section~\ref{sec:optim}, when the latent space $\mathcal{Z}$ is high-dimensional, the physically plausible and valid shapes usually live on a much-lower dimensional manifold $\cM\subset\mathcal{Z}$~\cite{Fefferman13a, Farghly25a}.
When using a gradient-based method and taking optimization steps of the form 
\begin{equation}
    \label{eq:gradient_step}
    \mathbf{z}^{(k+1)} = \mathbf{z}^{(k)} - \eta \nabla_{\mathbf{z}} \mathcal{O}\big(\mathcal{D}(\mathbf{z}^{(k)})\big) \; ,
\end{equation}
with $\eta$ a step size, the task gradient $\nabla_{\mathbf{z}} \mathcal{O}$ is computed solely to minimize the objective and $\mathbf{z}^{(k)}$ may drift further away from $\cM$ with each new iteration.
The pre-trained decoder $\mathcal{D}$ being trained on elements of $\cM$, the predictions risk gradually becoming less and less meaningful, devolving into adversarial or unrealistic shapes that numerically minimize the objective $\mathcal{O}$ without representing a valid 3D structure.

\subsection{Flow-corrected shape optimization}

\begin{figure}[t]
    \centering
    \begin{minipage}{0.49\textwidth}
        \centering
        \includegraphics[width=\linewidth]{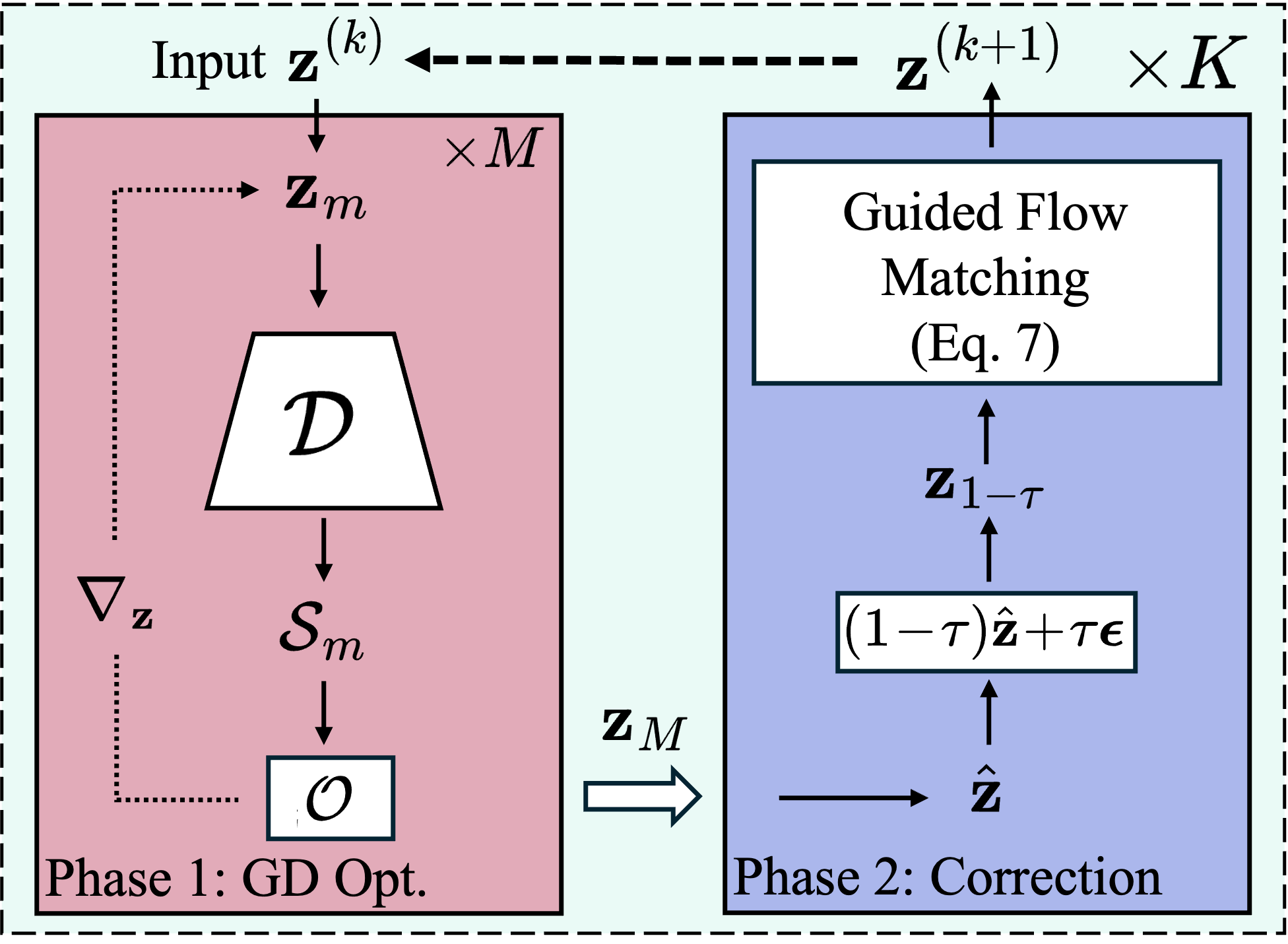}
        \caption{Method Diagram Overview. We alternate between gradient-based optimization (Phase 1) and correction via guided Flow Matching (Phase 2) to safely navigate the shape manifold and optimize the objective $\mathcal{O}$.}
        \label{fig:method}
    \end{minipage}
    \hfill %
    \begin{minipage}{0.5\textwidth}
        \vspace{-15pt}
        \begin{algorithm}[H] %
            \caption{Flow-Corrected Shape Optimization}
            \label{alg:algo}
            \small
            \begin{algorithmic}[1]
                \Require Initial $\mathbf{z}^{(0)}$, Objective $\mathcal{O}$
                \Require Flow Model $\mathbf{v}_\theta$, Decoder $\mathcal{D}$
                \Require Optimization iterations $K$, gradient steps $M$, step size $\eta$, noise ratio $\tau$, flow steps $N$

                \For{$k = 0, \dots, K-1$}
                    \State \textbf{Phase 1: Gradient-Based Minimization}
                    \State $\mathbf{z}_0 \gets \mathbf{z}^{(k)}$
                    \For{$m = 0, \dots, M-1$}
                        \State $\mathbf{z}_{m+1} \gets \mathbf{z}_m - \eta \nabla_{\mathbf{z}} \mathcal{O}\big(\mathcal{D}(\mathbf{z}_m)\big)$ (Eq.~\ref{eq:gradient_step})
                    \EndFor
                    \State $\hat{\mathbf{z}} \gets \mathbf{z}_M$, $c \gets \mathcal{O}\big(\mathcal{D}(\hat{\mathbf{z}})\big)$
                    
                    \State \textbf{Phase 2: Flow-Based Correction}
                    \State $\mathbf{x} \gets (1 - \tau)\hat{\mathbf{z}} + \tau \boldsymbol{\epsilon}$  (Eq.~\ref{eq:forward_noising})
                    \For{$j = 0, \dots, N-1$}  (Solve Eq.~\ref{eq:forward_ode})
                        \State $t_j \gets (1-\tau) + j \frac{\tau}{N}$
                        \State $\mathbf{x} \gets \mathbf{x} + \Delta t \cdot \tilde{\mathbf{v}}_\theta(\mathbf{x}, t_j, c)$ (Eq.~\ref{eq:cfg_guidance} or Eq.~\ref{eq:grad_guidance})
                    \EndFor
                    \State $\mathbf{z}^{(k+1)} \gets \mathbf{x}$
                \EndFor
                \State \Return $\mathbf{z}^{(K)}$
            \end{algorithmic}
        \end{algorithm}
    \end{minipage}
    \vspace{-2mm}
\end{figure}

To prevent this, we propose using a pre-trained Flow Matching model to keep the optimization trajectory close to the manifold.
In particular, we argue that alternating between direct gradient-descent optimization and flow-based correction, as outlined in Fig.~\ref{fig:method} and Algorithm~\ref{alg:algo}, allows a more robust optimization, which we demonstrate in our experiment.

\parag{Phase 1: Gradient-Based Minimization.} 
Given the current latent state $\mathbf{z}^{(k)}$ at optimization cycle $k$, we first take $M$ gradient steps (Eq.~\ref{eq:gradient_step}) to minimize $\mathcal{O}$ with respect to $\bz$.
After $M$ steps, this yields a potentially off-manifold latent vector $\hat{\bz}^{(k)}$.
This phase is the primary driver of the optimization to minimize the target objective.

\parag{Phase 2: Flow-Based Correction.}

To pull $\hat{\bz}^{(k)}$ back onto the valid manifold $\mathcal{M}$, we use a Flow Matching model $\mathbf{v}_\theta$ pre-trained on this manifold of the shape prior's latent space.

For this, we first perform a partial noising step and then exploit the straight-line probability path of the Flow Matching framework.
For a chosen noise injection ratio $\tau \in (0, 1)$, we directly interpolate between the latent code $\hat{\bz}^{(k)}$ and sampled noise $\boldsymbol{\epsilon} \sim \mathcal{N}(\mathbf{0}, \mathbf{I})$. This yields 
\begin{equation}
    \label{eq:forward_noising}
    \mathbf{z}_{1-\tau}^{(k)} = (1 - \tau)\hat{\bz}^{(k)} + \tau \boldsymbol{\epsilon} \; ,
\end{equation}
which pulls $\hat{\bz}^{(k)}$ towards the Gaussian noise, thereby attenuating the high-frequency adversarial artifacts that may be introduced by direct gradient descent.

Directly using the FM network $\mathbf{v}_\theta$  of Sec.~\ref{sec:preliminaries} for the correction towards the data manifold tends to undo the improvement brought about by gradient descent. Instead, we guide the flow field with the objective value attained at the end of the gradient phase, that is explicitly $c=\mathcal{O}\big(\mathcal{D}(\hat{\mathbf{z}}^{(k)})\big)$.
See Appendix~\ref{sec:appendix_ablation_guidance} for an ablation study showing that guidance helps achieve faster convergence.
This yields a guided vector field $\tilde{\mathbf{v}}_\theta$ that takes as input the current noised latent $\mathbf{z}_t$, the time $t$, and this condition $c$. It is computed using one of the two following ways. For notational simplicity, we omit the optimization cycle index $(k)$.

\hspace{5mm} \textbf{Option A: Classifier-Free Guidance (CFG)~\cite{Ho22a}.} 

The flow model $\mathbf{v}_\theta$ accepts an optional condition $c$ as additional input. During training, $c$ is set as the objective value of the current target shape $\mathcal{O}\big(\mathcal{D}(\mathbf{z}_1)\big)$. We use either the condition $c$ defined above or no condition denoted by $\emptyset$ to compute
\begin{equation}
    \label{eq:cfg_guidance}
    \tilde{\mathbf{v}}_\theta(\mathbf{z}_t, t, c) = \mathbf{v}_\theta(\mathbf{z}_t, t, \emptyset) + \omega \Big( \mathbf{v}_\theta(\mathbf{z}_t, t, c) - \mathbf{v}_\theta(\mathbf{z}_t, t, \emptyset) \Big) \; ,
\end{equation}
where $\omega > 1$ dictates the guidance strength.

\hspace{5mm} \textbf{Option B: Gradient-based Guidance.} 
The guided vector field is dynamically adjusted at each step to minimize the discrepancy between the current and target objective scores. We write
\begin{equation}
    \label{eq:grad_guidance}
    \tilde{\mathbf{v}}_\theta(\mathbf{z}_t, t, c)= \mathbf{v}_\theta(\mathbf{z}_t, t) - \lambda \nabla_{\mathbf{z}_t} \left( \mathcal{O}\big(\mathcal{D}(\mathbf{z}_t)\big) - c \right)^2 \; ,
\end{equation}
where $\lambda> 0$ is the guidance scale.

Option A is the preferred approach, as it leverages classifier-free guidance for faster execution. However, it requires the flow model to be conditioned on the optimization objective, which is not always possible. In contrast, option B leverages pre-trained models and is designed for large-scale approaches where retraining is impractical.

We integrate the ODE forward from $t = 1 - \tau$ to $t=1$, as in standard FM we write:
\begin{equation}
    \label{eq:forward_ode}
    \mathbf{z}^{(k+1)} = \mathbf{z}_{1-\tau}^{(k)} + \int_{1-\tau}^{1} \tilde{\mathbf{v}}_\theta(\mathbf{z}_t^{(k)}, t, c) \, dt \, .
\end{equation}
In practice, this continuous integral is approximated using a standard numerical ODE solver. 
In Appendix~\ref{app:theoretical_justification}, we leverage established flow theory to show why Phase 2 corrects manifold drift while maintaining optimization progress.

As discussed above and shown in the experiments, pure guided flow matching is not enough to effectively drive the optimization. Therefore, we propose this alternating strategy with gradient-descent.
This process is repeated for $K$ cycles, allowing the shape to progressively optimize the objective $\mathcal{O}$ while safely traveling along the valid shape manifold.
To prevent accumulated momentum from interfering with the updated latent trajectory, the optimizer is completely reset after every flow-matching correction step.


\section{Results}
\label{sec:result}

To demonstrate the efficacy and scalability of our proposed approach to complex models, we test it on several representative tasks and against state-of-the-art baselines.
We discuss our experimental setup below and then present our comparative results.

\subsection{Experimental setup}

Currently, no established benchmarks exist to evaluate 3D shape optimization within latent spaces.
We designed a comprehensive set of tasks, models, and metrics aimed at a fair and extensive comparison.
This benchmark will be released at \url{https://github.com/eseiler18/fcso}.

\parag{Tasks.}

We design three tasks of increasing complexity to evaluate our method.
The first involves reducing the volume of a given chair, which can be seen as a simple emulation of trying to make an object lighter. This is a simple objective that can be evaluated directly from the shape itself, yet can easily cause manifold drift by rewarding the partial disappearance of the object. Therefore, it serves as a simple yet useful testbed to demonstrate manifold drift.
The second, both more realistic and more complex, consists in refining the shape of a car to minimize its aerodynamic drag as evaluated by a surrogate model.
This is a meaningful but challenging engineering task because surrogate models are highly non-linear and often yield nonsensical results for shapes that are too far from the training examples, magnifying the manifold drift that we aim to tame.
The third experiment demonstrates scalability to the most recent and complex state-of-the-art models.
We work in the latent space of Hunyuan3D~\cite{Yang25a} and minimize the displacements under load by minimizing the {\it compliance}, as defined in~\cite{Zhan25a}, which can be understood as maximizing the stiffness of the objects.
Finally, to show that our method can extend beyond physics-based objectives, we evaluate an additional 2D silhouette-matching objective via differentiable rendering; its results are provided in Appendix~\ref{app:silhouette}.

\parag{Baselines.}

We compare our approach against several methods: standard Gradient Descent (GD)~\cite{Guillard24a,Zhan25a} and Flow Matching Guidance (FMG), either gradient-based or classifier-free~\cite{Ho22a}.
We also compare to D-Flow~\cite{BenHamu24}, which optimizes the starting point by differentiating through the entire flow trajectory; to ICTM~\cite{Zhang25e}, which enhances guidance by incorporating the score function and the trace of the velocity field's Jacobian; and to SGO~\cite{Chen26a}, a score-based regularization approach during gradient descent optimization.
Since our optimization starts from an initial shape rather than pure noise, we rely on an SDEdit-style~\cite{Meng22c} process for FMG and ICTM to enable editing rather than pure generation. 
Further details are provided in Appendix~\ref{app:baseline}.

\parag{Metrics.}

Beyond minimizing the objective, optimized shapes must remain meaningful.
To quantify this, we employ metrics commonly used to assess generative capabilities~\cite{Xiang25a, Zhang23d}, specifically computing the Fréchet Inception Distance~\cite{Heusel18a} and Kernel Inception Distance~\cite{Binkowski21a} between optimized shapes and a reference distribution of test shapes.
Distances are computed using 3D latent features from PointBert~\cite{Liu23i, Yu22d} and 2D features from 10 rendered views extracted via Inception~\cite{Szegedy15} and DINOv2~\cite{Oquab23}.
For a fair comparison, we evaluate all methods at matched optimization progress: each shape is assessed once its objective has been reduced by a fixed relative amount from its initial value (e.g., $50\%$ of initial volume). This ensures that differences in shape quality reflect the optimizer's behavior rather than how far it has driven the objective.
Additionally, to quantify how much the shape identity is preserved during optimization, we measure the perceptual similarity between the initial and optimized 3D shapes using the LPIPS metric~\cite{Zhang18h} applied to 2D renderings.
Details are provided in Appendix~\ref{app:metric}.

\subsection{Chair volume reduction}
\label{sec:chair}

\begin{figure}[t]
  \centering
  \includegraphics[width=0.9\textwidth]{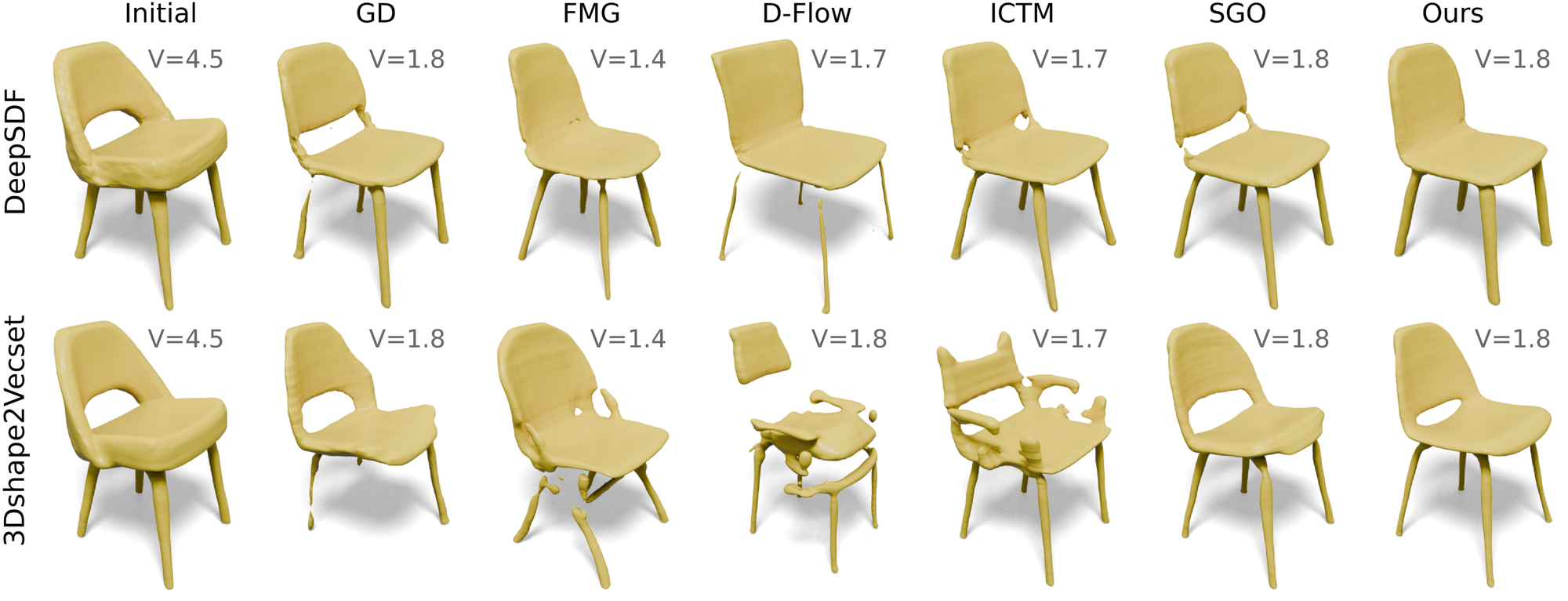} 
\caption{
Qualitative comparison of chair optimization using DeepSDF and 3DShape2VecSet with our method and baselines. 
$V$ denotes the volume. More examples are provided in Appendix~\ref{app:chair_results}}
\label{fig:chair_shape}
\vspace{-2mm} 
\end{figure}

Given a pre-trained decoder that generates chairs, our goal is to minimize the shape's volume while keeping them realistic, as shown in Fig.~\ref{fig:chair_shape}.
To demonstrate how all methods behave when the shape model increases in complexity, we use two: the older and simpler DeepSDF~\cite{Park19c} that decodes a single 256-D latent vector with an MLP and the more recent and complex 3DShape2VecSet~\cite{Zhang23d} that decodes a set of 256 vectors of size 8 with a Transformer.
We train them and volume-conditioned Flow Matching models on 2,419 watertight ShapeNet chairs~\cite{Chang15}. We then use an SDF-based approximation to compute and minimize the volume of 100 test chairs, as described in Appendix~\ref{app:volume_est}.
Given this conditioning, we use Classifier Free Guidance (Eq.~\ref{eq:cfg_guidance}) for both our method and FMG.
Additional implementation details are given in Appendix~\ref{app:chair}.
In Tab.~\ref{tab:chair_volume}, we report our quantitative results at 70\% volume reduction for all baselines. We provide more results in Appendix~\ref{app:chair_results}.


\begin{table}[t]
    \centering
    \caption{
    Chair volume optimization metrics at 70\% of volume reduction.
    Metrics are scaled for readability: $\text{KD}_{\text{incep}}$ ($\times 10^3$), $\text{FD}_{\text{point}}$ ($\times 10^1$), and $\text{KD}_{\text{point}}$ ($\times 10^4$).
    }
    \label{tab:chair_volume}
    \vspace{2mm}
    \small
    \setlength{\tabcolsep}{2.1pt}
    \resizebox{\textwidth}{!}{

    \begin{tabular}{l | c c c c c c | c c c c c c}
        \toprule
        \multirow{2}{*}{\textbf{Method}} & \multicolumn{6}{c|}{\textbf{DeepSDF}} & \multicolumn{6}{c}{\textbf{3DShape2VecSet}} \\
        & \textbf{FD}$_{\text{incep}}$ & \textbf{KD}$_{\text{incep}}$ & \textbf{FD}$_{\text{dino}}$ & \textbf{KD}$_{\text{dino}}$ & \textbf{FD}$_{\text{point}}$ & \textbf{KD}$_{\text{point}}$ 
        & \textbf{FD}$_{\text{incep}}$ & \textbf{KD}$_{\text{incep}}$ & \textbf{FD}$_{\text{dino}}$ & \textbf{KD}$_{\text{dino}}$ & \textbf{FD}$_{\text{point}}$ & \textbf{KD}$_{\text{point}}$ \\ \midrule
        GD & 47.37 & 13.00 & 368.42 & 19.51 & 2.79 & 3.11 & 88.12 & 55.63 & 659.34 & 32.43 & 3.94 & 6.44 \\
        FMG & 43.67 & 11.47 & 351.58 & 19.66 & 2.66 & 3.82 & 43.14 & 14.25 & 333.72 & 15.18 & 2.72 & 5.02 \\
        D-Flow & \textbf{36.29} & 10.28 & \textbf{334.13} & 21.29 & \textbf{2.36} & \textbf{2.88} & 56.92 & 23.45 & 426.55 & 19.01 & 3.24 & 4.81 \\
        ICTM & 43.65 & 10.06 & 351.45 & \textbf{16.78} & 2.62 & 3.19 & 91.71 & 12.90 & 576.96 & 15.22 & 3.89 & 3.25 \\
        SGO & 42.08 & 10.27 & 339.18 & 19.43 & 2.80 & 3.97 & 32.73 & 5.68 & 240.94 & 11.27 & 2.20 & 1.79 \\
        \textbf{Ours} & 41.12 & \textbf{9.72} & 335.42 & 19.66 & 2.49 & 3.00 & \textbf{31.76} & \textbf{4.83} & \textbf{234.61} & \textbf{10.66} & \textbf{1.95} & \textbf{1.42} \\
        \bottomrule
    \end{tabular} 
    }
\end{table}

As evidenced by Tab.~\ref{tab:chair_volume}, when using the smaller DeepSDF model, our approach operates on par with the baselines and brings no particular benefit.
This was to be expected: As DeepSDF operates in a small, compact latent space, it does not suffer from severe manifold drift. In contrast, the much more expressive 3DShape2VecSet operates on a much sparser higher-dimensional manifold where manifold drift has much more severe consequences and our approach comes into its own.
Here, we clearly outperform the other methods, yielding much more realistic chairs as shown in Fig.~\ref{fig:chair_shape}. An exception is SGO~\cite{Chen26a} that is a close second. However, that does not generalize to more complex tasks, as shown in the following experiments.
This highlights our method's ability to effectively leverage the expressivity of sophisticated models without suffering from manifold drift, whereas baselines struggle with the intricacies of navigating a more complex latent space. 
This trend is confirmed in the following experiments with more complex cases.

\subsection{Car drag optimization}
\label{sec:car}

Moving to a true engineering task, we now seek to minimize the aerodynamic drag ($C_d$) of a car.
We focus this study on the 3DShape2VecSet~\cite{Zhang23d} prior, for its higher capacity, and compare at two different levels of optimization progress.
We train this shape prior and a drag-conditioned Flow Matching model on 836 watertight ShapeNet cars~\cite{Chang15} with corresponding Computational Fluid Dynamics (CFD) simulation using OpenFOAM~\cite{OpenFoam}.
In theory, Adjoint Differentiation~\cite{Allaire15, Gao17a} can be used to compute $C_d$ in a differentiable manner.
However, this is extremely expensive because it requires a new simulation to be run each time~\cite{Alexandersen16} and can be complex to deploy.
Thus, following standard engineering practice~\cite{Jeong05,Laurenceau10,March11,Toal11,Xu17}, we use instead a surrogate model, specifically the GraphSAGE GCNN~\cite{Hamilton17, Bonnet22}.
Afterwards, we optimize 100 test cars to minimize their drag. We report quantitative results in Tab.~\ref{tab:drag_optimization_fixed} and qualitative ones in Fig.~\ref{fig:car_shape}.
As above, we use CFG (Eq.~\ref{eq:cfg_guidance}) for ours and for FMG. See Appendix~\ref{app:car_setup} for implementation details and Appendix~\ref{app:car_results} for additional results.


\begin{table}[t]
    \centering
    \caption{
        Car drag optimization (3DShape2VecSet) at 10\% and 30\% drag reduction.
        Metrics are scaled for readability: $\text{KD}_{\text{incep}}$ ($\times 10^3$), $\text{FD}_{\text{dino}}$ ($\times 10^{-1}$), $\text{FD}_{\text{point}}$ ($\times 10^2$), and $\text{KD}_{\text{point}}$ ($\times 10^5$).
    }
    \label{tab:drag_optimization_fixed}
    \vspace{2mm}
    \small
    \setlength{\tabcolsep}{2.1pt}
    \resizebox{\textwidth}{!}{
    \begin{tabular}{l | c c c c c c | c c c c c c}
        \toprule
        \multirow{2}{*}{\textbf{Method}} & \multicolumn{6}{c|}{\textbf{Drag 10\%}} & \multicolumn{6}{c}{\textbf{Drag 30\%}} \\
        & \textbf{FD}$_{\text{incep}}$ & \textbf{KD}$_{\text{incep}}$ & \textbf{FD}$_{\text{dino}}$ & \textbf{KD}$_{\text{dino}}$ & \textbf{FD}$_{\text{point}}$ & \textbf{KD}$_{\text{point}}$ 
        & \textbf{FD}$_{\text{incep}}$ & \textbf{KD}$_{\text{incep}}$ & \textbf{FD}$_{\text{dino}}$ & \textbf{KD}$_{\text{dino}}$ & \textbf{FD}$_{\text{point}}$ & \textbf{KD}$_{\text{point}}$ \\ \midrule
        GD     & 84.88 & 56.77 & 65.33 & 49.60 & 16.89 & 32.40 & 132.82 & 106.26 & 97.97 & 75.63 & 37.50 & 123.20 \\
        FMG    & 83.43 & 42.62 & 62.54 & 43.62 & 17.62 & 20.40 & 100.20 & 67.02  & 77.05 & 56.52 & 25.00 & 73.00  \\
        D-Flow & 72.33 & 42.74 & 61.76 & 45.86 & 15.38 & 25.20 & 108.79 & 76.44  & 82.85 & 62.68 & 29.95 & 76.70  \\
        ICTM   & 74.87 & 47.25 & 60.58 & 46.23 & 15.71 & 30.20 & 101.73 & 75.52  & 75.71 & 58.31 & 24.97 & 77.60  \\
        SGO    & 78.26 & 49.37 & 63.53 & 47.49 & 16.58 & 27.80 & 107.72 & 76.89  & 81.54 & 61.25 & 30.75 & 83.70  \\
        \textbf{Ours} & \textbf{62.42} & \textbf{34.81} & \textbf{52.79} & \textbf{40.68} & \textbf{11.35} & \textbf{12.80} & \textbf{89.08} & \textbf{55.91} & \textbf{69.20} & \textbf{52.72} & \textbf{20.96} & \textbf{51.40} \\
        \bottomrule
    \end{tabular} 
    }
\end{table}

\begin{figure}[t]
  \centering
  \includegraphics[width=\textwidth]{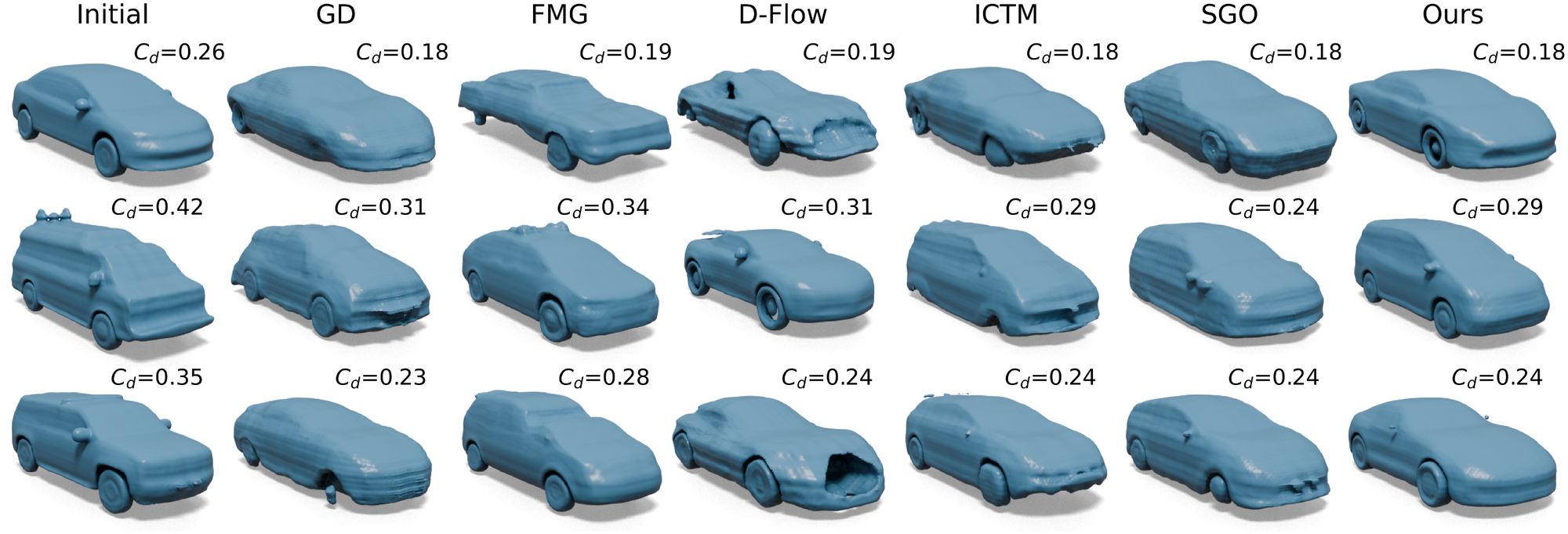} 
\caption{
Qualitative comparison of drag optimization. $C_d$ is the drag predicted by the surrogate. See Appendix~\ref{app:car_results} for more examples.
}
\label{fig:car_shape}
\end{figure}

Our approach outperforms the baselines on all metrics, with the gap increasing the further the optimization progress (10\% vs.\ 30\% drag reduction targets). It can be seen in Fig.~\ref{fig:car_shape} that the baselines exhibit manifold drift with over-smoothing, surface artifacts, and disappearing wheels.
The SGO results were relatively close to ours in the simpler volume reduction task of the previous section but the difference is much starker here. This is presumably due to the surrogate being highly non-linear with complex gradients and SGO's regularization, based on a stochastic score approximation, fails to maintain geometric validity and results in over-smoothing. 
We also confirm that these optimizations effectively minimize true physical drag by running simulations~\cite{OpenFoam}, see Appendix~\ref{app:simu}. 

\subsection{Minimizing displacements under load in Hunyuan3D}
\label{sec:phyiopt}

To show that our approach scales to truly complex state-of-the-art shape models, we test it on Hunyuan3D~\cite{Yang25a}.
It is a foundation model for high-fidelity 3D generation. It encodes shapes in a set of vectors of size $4096 \times 64$, which enables the capture of intricate geometric details. 
Unlike the category-specific priors used in our previous experiments, Hunyuan3D acts as a large-scale general prior trained on a vast range of objects.
While such a high-dimensional space is normally reserved for generation, we show that our method makes it possible to effectively exploit it for optimization.

\begin{figure}[t]
  \centering
  \includegraphics[width=\textwidth]{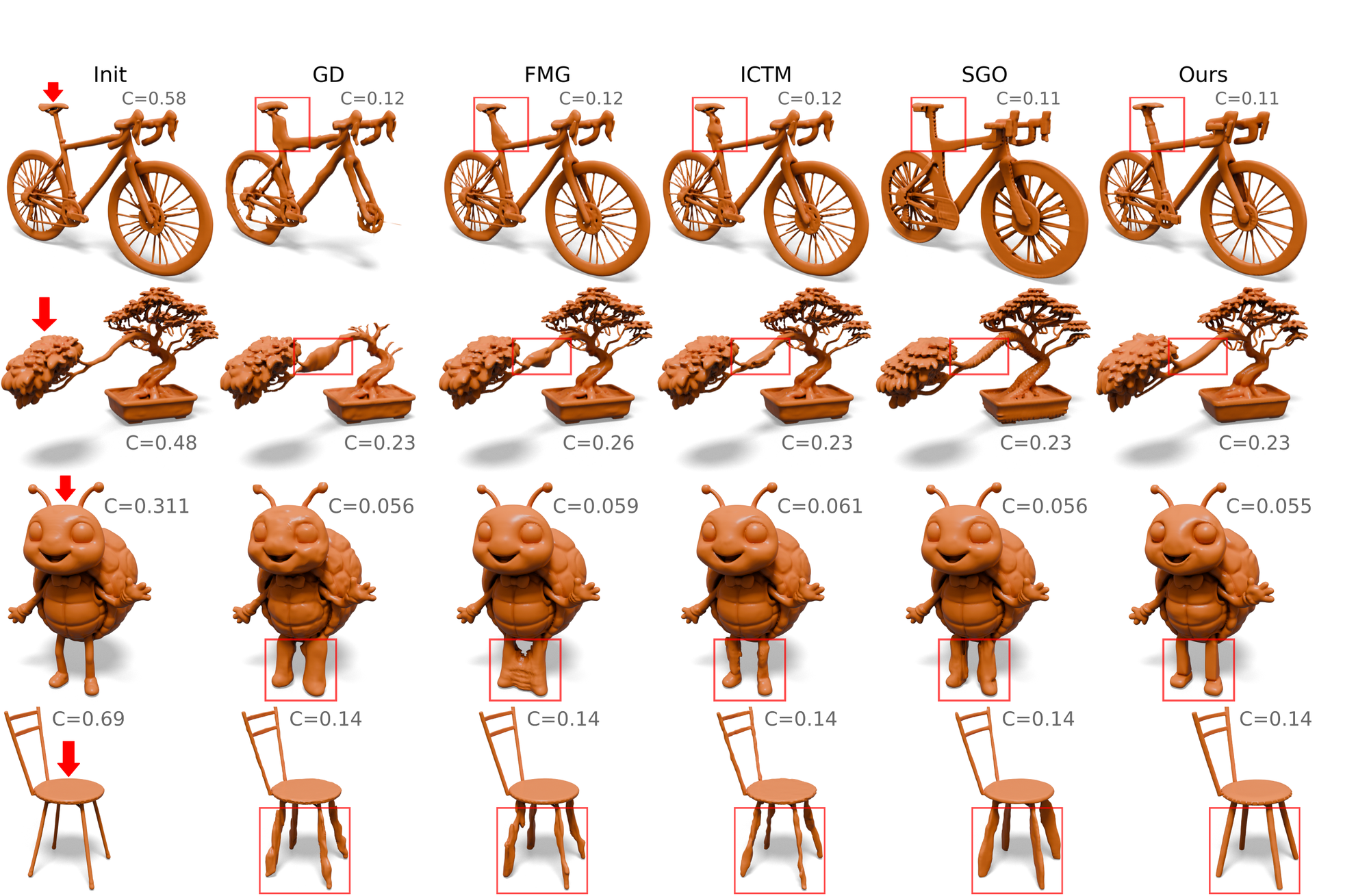} 
\caption{
We show the initial shapes with the applied force (red arrow) and optimized shapes across methods.
The compliance ($C$) is reported, where lower values indicate better structural stability.
}
\label{fig:phyiopt_shape}
\end{figure}


\begin{table}[htbp]
    \vspace{-4mm}
    \centering
    \caption{
    Compliance ($C$) optimization on BIKED bicycles and ABO chairs at 85\% and 65\% reduction, respectively.
    Scaling: $\text{KD}_{\text{incep}}$($\times10^3$), $\text{FD}_{\text{point}}$($\times 10^2$), $\text{KD}_{\text{point}}$($\times 10^5$).
    Examples in Appendix~\ref{app:Hunyuan3D_results}.
    }
    \label{tab:phyiopt_table_result}
    \vspace{2mm}

    \small
    \setlength{\tabcolsep}{2.1pt}
    \resizebox{\textwidth}{!}{
    \begin{tabular}{l | c c c c c c | c c c c c c}
        \toprule
        \multirow{2}{*}{\textbf{Method}} & \multicolumn{6}{c|}{\textbf{Bicycles}} & \multicolumn{6}{c}{\textbf{Chairs}} \\
        & \textbf{FD}$_{\text{incep}}$ & \textbf{KD}$_{\text{incep}}$ & \textbf{FD}$_{\text{dino}}$ & \textbf{KD}$_{\text{dino}}$ & \textbf{FD}$_{\text{point}}$ & \textbf{KD}$_{\text{point}}$
        & \textbf{FD}$_{\text{incep}}$ & \textbf{KD}$_{\text{incep}}$ & \textbf{FD}$_{\text{dino}}$ & \textbf{KD}$_{\text{dino}}$ & \textbf{FD}$_{\text{point}}$ & \textbf{KD}$_{\text{point}}$ \\ \midrule
        GD            & 43.09 & 19.39 & 267.63 & 19.71 & 20.84 & 28.82 & 68.18 & 29.69 & 575.84 & 39.67 & 32.71 & 27.80 \\
        FMG      & 47.34 & 15.17 & 263.37 & 12.59 & 22.21 & 21.51 & 55.74 & 19.45 & 431.28 & 26.57 & 27.21 & 16.00 \\
        D-Flow             & \multicolumn{6}{c|}{\textit{OOM}} & \multicolumn{6}{c}{\textit{OOM}} \\
        ICTM     & 47.41 & 14.87 & 278.98 & 12.61 & 23.03 & 25.63 & 55.92 & 19.37 & 432.10 & 26.80 & 28.05 & 18.33 \\
        SGO                & 21.13 & 1.54 & 84.32 & 1.08 & 12.97 & 4.48 & 55.08 & 16.20 & 426.41 & 22.49 & 29.31 & 12.40 \\
        \textbf{Ours}      & \textbf{20.65} & \textbf{1.12} & \textbf{82.37} & \textbf{0.81} & \textbf{12.41} & \textbf{0.38} & \textbf{41.27} & \textbf{9.22} & \textbf{282.41} & \textbf{15.89} & \textbf{23.07} & \textbf{5.20} \\
        \bottomrule
    \end{tabular}
    }
\end{table}

We minimize the physics-driven compliance $C$ introduced in PhysiOpt~\cite{Zhan25a}, which is equivalent to maximizing structural stiffness. In practice, this ensures the shape deforms as little as possible under loads such as the weight applied to a chair's seat or a bicycle's saddle.
See details in Appendix~\ref{app:compliance}.
While foundation models like Hunyuan3D are trained on highly diverse datasets (e.g., Objaverse~\cite{Deitke23a}), automatically assigning physically meaningful boundary conditions to arbitrary complex geometries remains a challenge.
We therefore select the chairs from the ABO dataset~\cite{Collinsand21a} (see Appendix~\ref{app:Hunyuan3D_results} for Tables) and Bicycles from the BIKED dataset~\cite{Regenwetter22} to evaluate our method on complex topologies where physical constraints can be reliably automatically defined.
We report our quantitative results in Tab.~\ref{tab:phyiopt_table_result}. Fig.~\ref{fig:phyiopt_shape} features qualitative results on a diverse set of shapes.
Implementation details are given in Appendix~\ref{app:Hunyuan3D_setup} and more results in Appendix~\ref{app:Hunyuan3D_results}.

Our approach again outperforms all baselines across all metrics, demonstrating superior preservation of the original data distribution, and leads to cleaner shape edits compared to the artifacts produced by other methods.
Notably, D-Flow proved computationally infeasible causing Out-Of-Memory (OOM) errors (NVIDIA A100 GPU with 40GB VRAM) because Hunyuan3D's architecture is too large, hence the missing numbers in the table.
Standard GD causes widespread geometric distortion, while FMG and ICTM correctly localize modifications to stressed regions but generate unnatural, bumpy surfaces. 
SGO performs better but still yields blocky and bumpy deformations, such as the bicycle frame and bonsai branch.
In contrast, our method naturally thickens the targeted regions while maintaining smooth, high-fidelity geometry.

\subsection{Shape identity preservation}
\label{sec:lpips_identity}

As shown in Tab.~\ref{tab:lpips_combined}, FCSO matches or achieves the lowest divergence from the original shape across all tasks.
This confirms that our method effectively preserves visual identity, rather than simply exploiting the optimization landscape to minimize the objective.
Nevertheless, aggressive targets can still alter the style of the shape: in Fig.~\ref{fig:car_shape} (third row), the SUV adopts a more sedan-like profile. While our method preserves identity better than the baselines, there remains room for improvement.

\begin{table}[t]
    \centering
    \caption{Shape identity preservation measured by LPIPS (mean $\pm$ std). Lower values indicate better preservation of the original shape's visual identity.}
    \label{tab:lpips_combined}
    \small
    \setlength{\tabcolsep}{4pt}
    \resizebox{\textwidth}{!}{
    \begin{tabular}{l | c c | c c | c c}
        \toprule
        \multirow{2}{*}{\textbf{Method}} & \multicolumn{2}{c|}{\textbf{Chair Volume (Tab.~\ref{tab:chair_volume})}} & \multicolumn{2}{c|}{\textbf{Car Drag (Tab.~\ref{tab:drag_optimization_fixed})}} & \multicolumn{2}{c}{\textbf{Hunyuan3D Compliance (Tab.~\ref{tab:phyiopt_table_result})}} \\
        & DeepSDF & 3DShape2VecSet & 10\% & 30\% & Bicycles & Chairs \\ \midrule
        GD     & $0.26 \pm 0.05$ & $0.29 \pm 0.07$ & $0.07 \pm 0.02$ & $0.14 \pm 0.02$ & $0.07 \pm 0.05$ & $0.08 \pm 0.06$ \\
        FMG    & $0.28 \pm 0.06$ & $0.27 \pm 0.05$ & $0.06 \pm 0.03$ & $0.10 \pm 0.04$ & $\mathbf{0.04 \pm 0.03}$ & $0.07 \pm 0.04$ \\
        D-Flow & $0.33 \pm 0.08$ & $0.25 \pm 0.06$ & $0.07 \pm 0.04$ & $0.11 \pm 0.03$ & \multicolumn{2}{c}{\textit{OOM}} \\
        ICTM   & $0.27 \pm 0.05$ & $0.24 \pm 0.05$ & $0.06 \pm 0.02$ & $0.08 \pm 0.04$ & $0.05 \pm 0.03$ & $0.07 \pm 0.05$ \\
        SGO    & $\mathbf{0.25 \pm 0.05}$ & $\mathbf{0.22 \pm 0.04}$ & $0.04 \pm 0.02$ & $0.08 \pm 0.03$ & $0.05 \pm 0.03$ & $\mathbf{0.05 \pm 0.07}$ \\
        \textbf{Ours} & $\mathbf{0.25 \pm 0.05}$ & $\mathbf{0.22 \pm 0.03}$ & $\mathbf{0.03 \pm 0.02}$ & $\mathbf{0.06 \pm 0.03}$ & $\mathbf{0.04 \pm 0.02}$ & $\mathbf{0.05 \pm 0.02}$ \\
        \bottomrule
    \end{tabular} 
    }
\end{table}


\section{Limitations}
\label{sec:lim}

While more efficient than back-propagation through the entire flow matching process, our approach is more computationally demanding than standard gradient descent because each optimization cycle requires $N$ forward evaluations of the vector field during the correction phase (see Appendix \ref{app:comp_res}).
Additionally, under high optimization targets, our method can occasionally introduce minor geometric artifacts or alter the style of the shape (Sec.~\ref{sec:lpips_identity}). However, it consistently preserves the overall shape semantics (see Appendix \ref{app:fail}).
We leave sparse voxel latents~\cite{Xiang25a} to future work, as adding material there requires an inpainting step~\cite{Zhan25a} that dominates the geometric change.
Furthermore, the hyperparameters governing the process, specifically the number of gradient steps $M$ and the noise injection ratio $\tau$, are currently fixed.
Future work can focus on developing adaptive mechanisms, potentially by leveraging surrogate model uncertainty or real-time shape quality metrics to dynamically balance objective minimization and shape quality preservation.


\section{Conclusion}

We have proposed a mechanism that tames the manifold drift problem when optimizing 3D shapes parameterized by high-dimensional latent vectors for performance.
This means that we can now handle complex composite objects that require such models.
Our next step will be to also enforce consistency constraints between the different parts of such objects to guarantee that, when one part is deformed, the others are deformed in a way that maintains the coherence of the whole object.
In other words, we will move from the unconstrained optimization described in this paper to optimization under consistency constraints.

\begin{ack}
This work was funded in part by the Swiss National Science Foundation.
\end{ack}

\bibliographystyle{plain}
\bibliography{bib/string, bib/learning, bib/vision, bib/graphics, bib/geom, bib/cfd, bib/misc, bib/biomed}

\newpage

\section*{NeurIPS Paper Checklist}

\begin{enumerate}

\item {\bf Claims}
    \item[] Question: Do the main claims made in the abstract and introduction accurately reflect the paper's contributions and scope?
    \item[] Answer: \answerYes{}
    \item[] Justification: Our contributions are clearly stated in the abstract and listed at the end of the introduction (Section \ref{sec:intro}). These claims are fully validated by the experimental results presented in Section \ref{sec:result}.
    \item[] Guidelines:
    \begin{itemize}
        \item The answer \answerNA{} means that the abstract and introduction do not include the claims made in the paper.
        \item The abstract and/or introduction should clearly state the claims made, including the contributions made in the paper and important assumptions and limitations. A \answerNo{} or \answerNA{} answer to this question will not be perceived well by the reviewers. 
        \item The claims made should match theoretical and experimental results, and reflect how much the results can be expected to generalize to other settings. 
        \item It is fine to include aspirational goals as motivation as long as it is clear that these goals are not attained by the paper. 
    \end{itemize}

\item {\bf Limitations}
    \item[] Question: Does the paper discuss the limitations of the work performed by the authors?
    \item[] Answer: \answerYes{}
    \item[] Justification: Yes, the limitations of our work are explicitly discussed in Section \ref{sec:lim}.

    \item[] Guidelines:
    \begin{itemize}
        \item The answer \answerNA{} means that the paper has no limitation while the answer \answerNo{} means that the paper has limitations, but those are not discussed in the paper. 
        \item The authors are encouraged to create a separate ``Limitations'' section in their paper.
        \item The paper should point out any strong assumptions and how robust the results are to violations of these assumptions (e.g., independence assumptions, noiseless settings, model well-specification, asymptotic approximations only holding locally). The authors should reflect on how these assumptions might be violated in practice and what the implications would be.
        \item The authors should reflect on the scope of the claims made, e.g., if the approach was only tested on a few datasets or with a few runs. In general, empirical results often depend on implicit assumptions, which should be articulated.
        \item The authors should reflect on the factors that influence the performance of the approach. For example, a facial recognition algorithm may perform poorly when image resolution is low or images are taken in low lighting. Or a speech-to-text system might not be used reliably to provide closed captions for online lectures because it fails to handle technical jargon.
        \item The authors should discuss the computational efficiency of the proposed algorithms and how they scale with dataset size.
        \item If applicable, the authors should discuss possible limitations of their approach to address problems of privacy and fairness.
        \item While the authors might fear that complete honesty about limitations might be used by reviewers as grounds for rejection, a worse outcome might be that reviewers discover limitations that aren't acknowledged in the paper. The authors should use their best judgment and recognize that individual actions in favor of transparency play an important role in developing norms that preserve the integrity of the community. Reviewers will be specifically instructed to not penalize honesty concerning limitations.
    \end{itemize}

\item {\bf Theory assumptions and proofs}
    \item[] Question: For each theoretical result, does the paper provide the full set of assumptions and a complete (and correct) proof?
\item {\bf Theory assumptions and proofs}
    \item[] Question: For each theoretical result, does the paper provide the full set of assumptions and a complete (and correct) proof?
    \item[] Answer: \answerNA{}
    \item[] Justification: We don't claim new theoretical result
    \item[] Guidelines:
    \begin{itemize}
        \item The answer \answerNA{} means that the paper does not include theoretical results. 
        \item All the theorems, formulas, and proofs in the paper should be numbered and cross-referenced.
        \item All assumptions should be clearly stated or referenced in the statement of any theorems.
        \item The proofs can either appear in the main paper or the supplemental material, but if they appear in the supplemental material, the authors are encouraged to provide a short proof sketch to provide intuition. 
        \item Inversely, any informal proof provided in the core of the paper should be complemented by formal proofs provided in appendix or supplemental material.
        \item Theorems and Lemmas that the proof relies upon should be properly referenced. 
    \end{itemize}

    \item {\bf Experimental result reproducibility}
    \item[] Question: Does the paper fully disclose all the information needed to reproduce the main experimental results of the paper to the extent that it affects the main claims and/or conclusions of the paper (regardless of whether the code and data are provided or not)?
    \item[] Answer: \answerYes{}
    \item[] Justification: Yes, all necessary implementation details and experimental setups required to reproduce our work are provided in Appendices \ref{app:chair_setup}, \ref{app:car_setup}, and \ref{app:Hunyuan3D_setup}.
    \item[] Guidelines:
    \begin{itemize}
        \item The answer \answerNA{} means that the paper does not include experiments.
        \item If the paper includes experiments, a \answerNo{} answer to this question will not be perceived well by the reviewers: Making the paper reproducible is important, regardless of whether the code and data are provided or not.
        \item If the contribution is a dataset and\slash or model, the authors should describe the steps taken to make their results reproducible or verifiable. 
        \item Depending on the contribution, reproducibility can be accomplished in various ways. For example, if the contribution is a novel architecture, describing the architecture fully might suffice, or if the contribution is a specific model and empirical evaluation, it may be necessary to either make it possible for others to replicate the model with the same dataset, or provide access to the model. In general. releasing code and data is often one good way to accomplish this, but reproducibility can also be provided via detailed instructions for how to replicate the results, access to a hosted model (e.g., in the case of a large language model), releasing of a model checkpoint, or other means that are appropriate to the research performed.
        \item While NeurIPS does not require releasing code, the conference does require all submissions to provide some reasonable avenue for reproducibility, which may depend on the nature of the contribution. For example
        \begin{enumerate}
            \item If the contribution is primarily a new algorithm, the paper should make it clear how to reproduce that algorithm.
            \item If the contribution is primarily a new model architecture, the paper should describe the architecture clearly and fully.
            \item If the contribution is a new model (e.g., a large language model), then there should either be a way to access this model for reproducing the results or a way to reproduce the model (e.g., with an open-source dataset or instructions for how to construct the dataset).
            \item We recognize that reproducibility may be tricky in some cases, in which case authors are welcome to describe the particular way they provide for reproducibility. In the case of closed-source models, it may be that access to the model is limited in some way (e.g., to registered users), but it should be possible for other researchers to have some path to reproducing or verifying the results.
        \end{enumerate}
    \end{itemize}

\item {\bf Open access to data and code}
    \item[] Question: Does the paper provide open access to the data and code, with sufficient instructions to faithfully reproduce the main experimental results, as described in supplemental material?
    \item[] Answer: \answerNo{}
    \item[] Justification: The code and benchmark will be released at \url{https://github.com/eseiler18/fcso}. In the meantime, comprehensive details, including pseudo-code \ref{alg:algo} and hyperparameters in Tables \ref{tab:chair_optim_hyperparams}, \ref{tab:car_optim_hyperparams}, \ref{tab:physiopt_optim_hyperparams}, are provided to allow independent reproduction.
    \item[] Guidelines:
    \begin{itemize}
        \item The answer \answerNA{} means that paper does not include experiments requiring code.
        \item Please see the NeurIPS code and data submission guidelines (\url{https://neurips.cc/public/guides/CodeSubmissionPolicy}) for more details.
        \item While we encourage the release of code and data, we understand that this might not be possible, so \answerNo{} is an acceptable answer. Papers cannot be rejected simply for not including code, unless this is central to the contribution (e.g., for a new open-source benchmark).
        \item The instructions should contain the exact command and environment needed to run to reproduce the results. See the NeurIPS code and data submission guidelines (\url{https://neurips.cc/public/guides/CodeSubmissionPolicy}) for more details.
        \item The authors should provide instructions on data access and preparation, including how to access the raw data, preprocessed data, intermediate data, and generated data, etc.
        \item The authors should provide scripts to reproduce all experimental results for the new proposed method and baselines. If only a subset of experiments are reproducible, they should state which ones are omitted from the script and why.
        \item At submission time, to preserve anonymity, the authors should release anonymized versions (if applicable).
        \item Providing as much information as possible in supplemental material (appended to the paper) is recommended, but including URLs to data and code is permitted.
    \end{itemize}

\item {\bf Experimental setting/details}
    \item[] Question: Does the paper specify all the training and test details (e.g., data splits, hyperparameters, how they were chosen, type of optimizer) necessary to understand the results?
    \item[] Answer: \answerYes{}
    \item[] Justification: Yes, all necessary implementation details and experimental setups required to reproduce our work are provided in Appendices \ref{app:chair_setup}, \ref{app:car_setup}, and \ref{app:Hunyuan3D_setup}.
    \item[] Guidelines:
    \begin{itemize}
        \item The answer \answerNA{} means that the paper does not include experiments.
        \item The experimental setting should be presented in the core of the paper to a level of detail that is necessary to appreciate the results and make sense of them.
        \item The full details can be provided either with the code, in appendix, or as supplemental material.
    \end{itemize}

\item {\bf Experiment statistical significance}
    \item[] Question: Does the paper report error bars suitably and correctly defined or other appropriate information about the statistical significance of the experiments?
    \item[] Answer: \answerNo{}
    \item[] Justification: Our primary metrics are FID and KID. As these are population-level metrics, computing error bars requires rerunning the entire optimization process for the full dataset multiple times. Given the high computational cost of 3D shape optimization, this was not performed. For the per-shape LPIPS metric, we report the mean and standard deviation across all optimized shapes (Tab.~\ref{tab:lpips_combined}).    \item[] Guidelines:
    \begin{itemize}
        \item The answer \answerNA{} means that the paper does not include experiments.
        \item The authors should answer \answerYes{} if the results are accompanied by error bars, confidence intervals, or statistical significance tests, at least for the experiments that support the main claims of the paper.
        \item The factors of variability that the error bars are capturing should be clearly stated (for example, train/test split, initialization, random drawing of some parameter, or overall run with given experimental conditions).
        \item The method for calculating the error bars should be explained (closed form formula, call to a library function, bootstrap, etc.)
        \item The assumptions made should be given (e.g., Normally distributed errors).
        \item It should be clear whether the error bar is the standard deviation or the standard error of the mean.
        \item It is OK to report 1-sigma error bars, but one should state it. The authors should preferably report a 2-sigma error bar than state that they have a 96\% CI, if the hypothesis of Normality of errors is not verified.
        \item For asymmetric distributions, the authors should be careful not to show in tables or figures symmetric error bars that would yield results that are out of range (e.g., negative error rates).
        \item If error bars are reported in tables or plots, the authors should explain in the text how they were calculated and reference the corresponding figures or tables in the text.
    \end{itemize}

\item {\bf Experiments compute resources}
    \item[] Question: For each experiment, does the paper provide sufficient information on the computer resources (type of compute workers, memory, time of execution) needed to reproduce the experiments?
    \item[] Answer: \answerYes{}
    \item[] Justification: Yes, full details regarding the hardware, memory configurations are provided in Appendix \ref{app:comp_res}.
    \item[] Guidelines:
    \begin{itemize}
        \item The answer \answerNA{} means that the paper does not include experiments.
        \item The paper should indicate the type of compute workers CPU or GPU, internal cluster, or cloud provider, including relevant memory and storage.
        \item The paper should provide the amount of compute required for each of the individual experimental runs as well as estimate the total compute. 
        \item The paper should disclose whether the full research project required more compute than the experiments reported in the paper (e.g., preliminary or failed experiments that didn't make it into the paper). 
    \end{itemize}
    
\item {\bf Code of ethics}
    \item[] Question: Does the research conducted in the paper conform, in every respect, with the NeurIPS Code of Ethics \url{https://neurips.cc/public/EthicsGuidelines}?
    \item[] Answer: \answerYes{}
    \item[] Justification: We have reviewed the NeurIPS Code of Ethics and confirm that this research strictly adheres to it in every respect.    
    \item[] Guidelines:
    \begin{itemize}
        \item The answer \answerNA{} means that the authors have not reviewed the NeurIPS Code of Ethics.
        \item If the authors answer \answerNo, they should explain the special circumstances that require a deviation from the Code of Ethics.
        \item The authors should make sure to preserve anonymity (e.g., if there is a special consideration due to laws or regulations in their jurisdiction).
    \end{itemize}

\item {\bf Broader impacts}
    \item[] Question: Does the paper discuss both potential positive societal impacts and negative societal impacts of the work performed?
    \item[] Answer: \answerNA{}
    \item[] Justification: This paper presents foundational algorithmic research on 3D latent space optimization. As a generic mathematical framework, it is not tied to specific deployments and does not have a direct path to foreseeable negative societal impacts.
    \item[] Guidelines: 
    \begin{itemize}
        \item The answer \answerNA{} means that there is no societal impact of the work performed.
        \item If the authors answer \answerNA{} or \answerNo, they should explain why their work has no societal impact or why the paper does not address societal impact.
        \item Examples of negative societal impacts include potential malicious or unintended uses (e.g., disinformation, generating fake profiles, surveillance), fairness considerations (e.g., deployment of technologies that could make decisions that unfairly impact specific groups), privacy considerations, and security considerations.
        \item The conference expects that many papers will be foundational research and not tied to particular applications, let alone deployments. However, if there is a direct path to any negative applications, the authors should point it out. For example, it is legitimate to point out that an improvement in the quality of generative models could be used to generate Deepfakes for disinformation. On the other hand, it is not needed to point out that a generic algorithm for optimizing neural networks could enable people to train models that generate Deepfakes faster.
        \item The authors should consider possible harms that could arise when the technology is being used as intended and functioning correctly, harms that could arise when the technology is being used as intended but gives incorrect results, and harms following from (intentional or unintentional) misuse of the technology.
        \item If there are negative societal impacts, the authors could also discuss possible mitigation strategies (e.g., gated release of models, providing defenses in addition to attacks, mechanisms for monitoring misuse, mechanisms to monitor how a system learns from feedback over time, improving the efficiency and accessibility of ML).
    \end{itemize}
    
\item {\bf Safeguards}
    \item[] Question: Does the paper describe safeguards that have been put in place for responsible release of data or models that have a high risk for misuse (e.g., pre-trained language models, image generators, or scraped datasets)?
    \item[] Answer: \answerNA{}
    \item[] Justification: Our paper introduces an optimization algorithm for 3D shapes rather than releasing a new high-risk foundational model or a scraped dataset. We rely on existing, standard public datasets (ShapeNet, ABO, BIKED) and pre-trained priors, which do not require additional safeguards for release.
    \item[] Guidelines:
    \begin{itemize}
        \item The answer \answerNA{} means that the paper poses no such risks.
        \item Released models that have a high risk for misuse or dual-use should be released with necessary safeguards to allow for controlled use of the model, for example by requiring that users adhere to usage guidelines or restrictions to access the model or implementing safety filters. 
        \item Datasets that have been scraped from the Internet could pose safety risks. The authors should describe how they avoided releasing unsafe images.
        \item We recognize that providing effective safeguards is challenging, and many papers do not require this, but we encourage authors to take this into account and make a best faith effort.
    \end{itemize}

\item {\bf Licenses for existing assets}
    \item[] Question: Are the creators or original owners of assets (e.g., code, data, models), used in the paper, properly credited and are the license and terms of use explicitly mentioned and properly respected?
    \item[] Answer: \answerYes{}
    \item[] Justification: Yes, all creators of the datasets (ShapeNet~\cite{Chang15}, ABO~\cite{Collinsand21a}, BIKED~\cite{Regenwetter22}) and models (DeepSDF~\cite{Park19c}, 3DShape2VecSet~\cite{Zhang23d}, Hunyuan3D~\cite{Yang25a}) used in our study are properly cite.
    \item[] Guidelines:
    \begin{itemize}
        \item The answer \answerNA{} means that the paper does not use existing assets.
        \item The authors should cite the original paper that produced the code package or dataset.
        \item The authors should state which version of the asset is used and, if possible, include a URL.
        \item The name of the license (e.g., CC-BY 4.0) should be included for each asset.
        \item For scraped data from a particular source (e.g., website), the copyright and terms of service of that source should be provided.
        \item If assets are released, the license, copyright information, and terms of use in the package should be provided. For popular datasets, \url{paperswithcode.com/datasets} has curated licenses for some datasets. Their licensing guide can help determine the license of a dataset.
        \item For existing datasets that are re-packaged, both the original license and the license of the derived asset (if it has changed) should be provided.
        \item If this information is not available online, the authors are encouraged to reach out to the asset's creators.
    \end{itemize}

\item {\bf New assets}
    \item[] Question: Are new assets introduced in the paper well documented and is the documentation provided alongside the assets?
  \item[] Answer: \answerNA{}
    \item[] Justification: We are not releasing the source code, new datasets, or new models at this time. Therefore, no new assets are introduced in this submission.
    \item[] Guidelines:
    \begin{itemize}
        \item The answer \answerNA{} means that the paper does not release new assets.
        \item Researchers should communicate the details of the dataset\slash code\slash model as part of their submissions via structured templates. This includes details about training, license, limitations, etc. 
        \item The paper should discuss whether and how consent was obtained from people whose asset is used.
        \item At submission time, remember to anonymize your assets (if applicable). You can either create an anonymized URL or include an anonymized zip file.
    \end{itemize}

\item {\bf Crowdsourcing and research with human subjects}
    \item[] Question: For crowdsourcing experiments and research with human subjects, does the paper include the full text of instructions given to participants and screenshots, if applicable, as well as details about compensation (if any)? 
    \item[] Answer: \answerNA{}
    \item[] Justification:  The paper does not involve crowdsourcing nor research with human subjects.
    \item[] Guidelines:
    \begin{itemize}
        \item The answer \answerNA{} means that the paper does not involve crowdsourcing nor research with human subjects.
        \item Including this information in the supplemental material is fine, but if the main contribution of the paper involves human subjects, then as much detail as possible should be included in the main paper. 
        \item According to the NeurIPS Code of Ethics, workers involved in data collection, curation, or other labor should be paid at least the minimum wage in the country of the data collector. 
    \end{itemize}

\item {\bf Institutional review board (IRB) approvals or equivalent for research with human subjects}
    \item[] Question: Does the paper describe potential risks incurred by study participants, whether such risks were disclosed to the subjects, and whether Institutional Review Board (IRB) approvals (or an equivalent approval/review based on the requirements of your country or institution) were obtained?
    \item[] Answer: \answerNA{}
    \item[] Justification: The paper does not involve crowdsourcing nor research with human subjects.
    \item[] Guidelines:
    \begin{itemize}
        \item The answer \answerNA{} means that the paper does not involve crowdsourcing nor research with human subjects.
        \item Depending on the country in which research is conducted, IRB approval (or equivalent) may be required for any human subjects research. If you obtained IRB approval, you should clearly state this in the paper. 
        \item We recognize that the procedures for this may vary significantly between institutions and locations, and we expect authors to adhere to the NeurIPS Code of Ethics and the guidelines for their institution. 
        \item For initial submissions, do not include any information that would break anonymity (if applicable), such as the institution conducting the review.
    \end{itemize}

\item {\bf Declaration of LLM usage}
    \item[] Question: Does the paper describe the usage of LLMs if it is an important, original, or non-standard component of the core methods in this research? Note that if the LLM is used only for writing, editing, or formatting purposes and does \emph{not} impact the core methodology, scientific rigor, or originality of the research, declaration is not required.
    \item[] Answer: \answerNA{}
    \item[] Justification: The core method development in this research does not involve LLMs,
    \item[] Guidelines:
    \begin{itemize}
        \item The answer \answerNA{} means that the core method development in this research does not involve LLMs as any important, original, or non-standard components.
        \item Please refer to our LLM policy in the NeurIPS handbook for what should or should not be described.
    \end{itemize}

\end{enumerate}

\appendix


\newpage
\section*{Appendix}
\addcontentsline{toc}{section}{Appendix}

\etocsettocstyle{\subsection*{Contents of the Appendix}}{}
\localtableofcontents

\renewcommand{\thesubsection}{\Alph{subsection}}

\newpage

\subsection{Metrics}
\label{app:metric}

To quantify the quality of the optimized shapes, we measure the distributional discrepancy between set of optimized shapes and a set of reference shapes.
For the chair (\ref{sec:chair}) and car (\ref{sec:car}) experiments, the reference set is constructed using ground-truth shapes from ShapeNet.
For the PhysiOpt optimization on the ABO and BIKED datasets (\ref{sec:phyiopt}), the reference set comprises shapes extracted via Hunyuan3D using the dataset's images.
We enforce strict separation: the reference set is computed entirely independently of the shapes that we optimize.
Furthermore, to guarantee a fair comparison across methods, we compute the set of optimized shapes at a fixed objective reduction threshold.
This allows us to quantify shape quality once all methods have reached an equivalent optimization state.

The metrics below are computed on feature embeddings extracted from 3D geometries using pre-trained PointBert \cite{Liu23i, Yu22d}, as well as from 10 multi-view 2D renders processed through pre-trained Inception \cite{Szegedy15} and DINOv2 \cite{Oquab23} networks.

\paragraph{Fréchet Distance}~\cite{Heusel18a}
This metric is most widely recognized by its seminal formulation, the Fréchet Inception Distance (FID).
Although we extract features using multiple backbones (PointBert, DINOv2) in addition to Inception, we retain the widely adopted FID nomenclature and mathematical formulation.
FID approximates the feature distributions as multivariate Gaussians and computes the Wasserstein-2 distance between them. Let $(\mu_r, \Sigma_r)$ and $(\mu_o, \Sigma_o)$ denote the mean and covariance of the features from the reference set and the optimized set, respectively.
The FID is defined as:\begin{equation}
    FID = \|\mu_r - \mu_o\|^2 + \text{Tr}\big(\Sigma_r + \Sigma_o - 2(\Sigma_r \Sigma_o)^{1/2}\big)
\end{equation}
where $\text{Tr}$ denotes the trace of a matrix. A lower FID indicates that the optimized shapes closely match the structural fidelity of the reference dataset.

\paragraph{Kernel Distance} \cite{Binkowski21a}
Following a similar convention, while this metric computes the kernel Maximum Mean Discrepancy (MMD), it is commonly referred to in the generative literature as the Kernel Inception Distance (KID).
We maintain this standard terminology across all our feature extractors. Unlike FID, KID does not assume a Gaussian distribution.
It computes the squared MMD between feature representations using a kernel function.
We also compute this metrics using multiple backbones (PointBert, DINOv2) in addition to Inception.
Given the feature distribution of the reference set $P_r$, the distribution of the optimized set $P_o$, and a kernel function $k(x, y)$, the KID is formulated as:\begin{equation}
    KID = \mathbb{E}_{x, x' \sim P_r}[k(x, x')] + \mathbb{E}_{y, y' \sim P_o}[k(y, y')] - 2\mathbb{E}_{x \sim P_r, y \sim P_o}[k(x, y)]
\end{equation}
We utilize a polynomial kernel $k(x, y) = (\frac{1}{d} x^T y + 1)^3$, where $d$ is the feature dimension. Lower KID scores confirm that the optimized geometries remain valid and structurally aligned with the reference shapes.

\paragraph{LPIPS} \cite{Zhang18h}
To quantify the preservation of shape identity during optimization, we employ the Learned Perceptual Image Patch Similarity (LPIPS) metric.
Unlike Fréchet or Kernel distances which measure the distance between population distributions, LPIPS computes the perceptual similarity between pairs of individual shapes (specifically, between the initial shape and its optimized counterpart). 
We extract multi-view 2D renderings of both the initial and optimized 3D shapes and compute the distance between their feature maps extracted from a pre-trained network (AlexNet).
By operating in feature space, LPIPS effectively captures whether the original visual identity and structure are maintained, while naturally accommodating the valid, physics-driven deformations required by the optimization.
A lower LPIPS score indicates that the optimized shape closely preserves the visual identity of the initial shape.

\subsection{Baselines}
\label{app:baseline}
In this section, we detail the baselines used for comparison and explain how we adapt them, when necessary, to our specific task of shape optimization.

\paragraph{Gradient Descent (GD)} 
GD serves as our most straightforward baseline, representing a standard approach widely employed in various shape optimization problems \cite{Guillard24a, Zhan25a}.
Beyond acting as a baseline, it also functions as an ablation study: omitting Phase 2 of our proposed method effectively reduces the entire pipeline to standard gradient descent.
To encourage the optimization to stay close to the manifold of valid shapes, we augment the objective with an $L_2$ regularization term anchored to the initial latent code, adding $\lambda \|z - z_{init}\|_2^2$ to the loss, where $z$ is the current latent vector, $z_{init}$ is the initial starting point, and $\lambda$ is the regularization weight.

\paragraph{Flow Matching Guidance (FMG)}
FMG serves as an additional baseline that also functions as an ablation study, as it isolates Phase 2 of our proposed pipeline by omitting the gradient descent step.
Adopting an SDEdit-style approach \cite{Meng22c}, we partially noise the latent representation and subsequently perform guided flow matching to optimize the shape, utilizing the guidance formulations defined in either Eq.~\ref{eq:cfg_guidance} or Eq.~\ref{eq:grad_guidance}.
Equation \ref{eq:cfg_guidance} was used for the chair volume (\ref{sec:chair}) and car drag (\ref{sec:car}) experiments, as the underlying flow matching model could be conditioned on the optimization objective during training.
For the Hunyuan3D experiment (\ref{sec:phyiopt}), where the model was unconditioned with respect to the optimization objective, we resorted to the gradient guidance formulation in Eq. \ref{eq:grad_guidance}.

\paragraph{D-Flow}
Originally designed to solve image inverse problems, D-Flow~\cite{BenHamu24} directly optimizes the initial noise vector that serves as the input to the flow matching process.
However, this approach is computationally expensive, as it requires back-propagating gradients through the entire ODE trajectory, from the generated data back to the initial noise distribution.
To adapt D-Flow to our shape optimization task, we initialize the optimization process using the specific noise vector that generates our starting shape.
This initial noise vector is pre-computed via an optimization-based latent inversion process prior to the shape optimization.

\paragraph{Iterative Corrupted Trajectory Matching (ICTM)}
Developed for linear inverse problems in image restoration, ICTM~\cite{Zhang25e} enhances guided flow matching through a local MAP objective that combines a score-based prior with a regularization term derived from the velocity field's Jacobian trace.
To adapt this framework to our shape optimization task, we employ an SDEdit-style approach~\cite{Meng22c} by initializing the generative trajectory from a partially noised latent vector rather than pure noise, thereby maintaining structural proximity to the initial shape.

\paragraph{Score Guided Optimization (SGO)} 
Score Guided Optimization (SGO) \cite{Chen26a} leverages a regularization term based on a stochastic score approximation, $-\nabla_{\mathbf{x}} \log p(\mathbf{x})$, which is controlled by a hyperparameter $\lambda$.
This $\lambda$ parameter is empirically tuned through trial and error to find the optimal balance between objective minimization and shape validity for each experiment.

\subsection{Computational resources and runtimes}
\label{app:comp_res}

All experiments were conducted on a GPU cluster. Specifically, the volume reduction experiments (Section \ref{sec:chair}) and the aerodynamic drag optimization experiments (Section \ref{sec:car}) were executed using NVIDIA V100 GPUs (32GB VRAM). Due to the higher dimensionality and larger memory footprint of the Hunyuan3D model, the compliance minimization experiments (Section \ref{sec:phyiopt}) required NVIDIA A100 GPUs (40GB VRAM).


\begin{table}[htbp]
    \centering
    \caption{Average computational cost per shape optimization run across different experimental setups and optimization methods. The objectives are a 70\% volume reduction for Chair (DeepSDF/VecSet), a 30\% drag reduction for Car (VecSet), a 65\% compliance reduction for Hunyuan Chair, and a 50\% compliance reduction for Hunyuan Table.}
    \label{tab:computational_resources_summary}
    \vspace{2mm}
    \small
    \setlength{\tabcolsep}{8pt} 
    \resizebox{\textwidth}{!}{
    \begin{tabular}{l | c c c c c}
        \toprule
        \textbf{Method} & \textbf{Chair (DeepSDF)} & \textbf{Chair (VecSet)} & \textbf{Car (VecSet)} & \textbf{Hunyuan Chair} & \textbf{Hunyuan Table} \\ \midrule
        GD     & 38s & 1m 02s & 53s & 3m 18s & 2m 34s \\
        FMG    & 10s & 14s & 15s & 55s & 59s \\
        D-Flow & 3m 51s & 14m 56s & 17m 10s & OOM & OOM \\
        ICTM   & 30s & 54s & 55s & 2m16s & 2m11s \\
        SGO    & 2m 10s & 3m 20s & 3m 55s & 6m 09s & 6m 58s \\
        Ours & 2m 15s & 3m 25s & 3m 12s & 6m 21s & 6m 21s \\
        \bottomrule
    \end{tabular} 
    }
\end{table}

Table~\ref{tab:computational_resources_summary} provides a detailed overview of the average computational cost across the different experimental setups.
FMG is as the most computationally efficient method, particularly for the first three experimental setups, as it only requires a single pass through the Flow Matching model (utilizing CFG).
ICTM requires slightly more time with the additional term apply during guidance.
Standard gradient descent (GD) takes comparatively longer. 
Furthermore, SGO and our proposed method exhibit higher runtimes than pure GD due to the additional overhead of evaluating the Flow Matching model.
Finally, D-Flow incurs the highest computational cost because it requires backpropagating through the entire ODE trajectory at each step.
This intensive memory requirement makes D-Flow completely infeasible for larger architectures like Hunyuan3D, resulting in Out-Of-Memory (OOM) errors (NVIDIA A100 GPU with 40GB VRAM).

\subsection{Theoretical insights: preventing manifold drift}
\label{app:theoretical_justification}

Recall from Section~\ref{sec:method} that our alternating optimization scheme aims to refine a latent vector $\bz$ to minimize a task-specific objective $\mathcal{O}(\mathcal{D}(\bz))$. Let $\hat{\bz}^{(k)}$ denote the intermediate latent vector obtained after the gradient-based minimization (Phase 1) at cycle $k$, achieving an objective value $c = \mathcal{O}\big(\mathcal{D}(\hat{\bz}^{(k)})\big)$. Furthermore, in our Flow Matching formulation, time $t \in [0,1]$ is defined such that $t=0$ corresponds to the standard Gaussian prior and $t=1$ corresponds to the clean data distribution.

Let $p_{\text{data}}(\bz)$ denote the probability density function of the shape prior's training data, which is also the distribution our Flow Matching model is trained to approximate.
Meaningful 3D shapes correspond to a high-probability region, such that $p_{\text{data}}(\bz) \approx 1$.
Conversely, adversarial or meaningless shapes reside in off-manifold regions where $p_{\text{data}}(\bz) \approx 0$.
During Phase 1, the gradient descent purely minimizes the objective $\mathcal{O}$ without accounting for this underlying data distribution.
Consequently, the intermediate latent vector can drift into low-probability regions, meaning $p_{\text{data}}(\hat{\bz}^{(k)}) \to 0$.

To formalize how Phase 2 corrects this drift and pulls the latent vector back to regions where $p_{\text{data}} \approx 1$, we analyze the continuous-time dynamics of our flow-based correction.
Under the Optimal Transport probability path, the noisy latent state follows $\bz_t | \bz_1 \sim \mathcal{N}(t\bz_1, (1-t)^2 \mathbf{I})$.
The exact marginal vector field driving this generative flow is given by:
\begin{equation}
    \mathbf{v}(\bz_t, t) = \frac{\mathbb{E}[\bz_1 | \bz_t] - \bz_t}{1-t} 
\end{equation}

Leveraging Tweedie's formula \cite{Efron11a}, we can express the posterior expectation of the uncorrupted data $\bz_1$ precisely through the score function of the smoothed marginal density $p_t(\bz_t)$:
\begin{equation}
    \mathbb{E}[t\bz_1 | \bz_t] = \bz_t + (1-t)^2 \nabla_{\bz_t} \log p_t(\bz_t) 
\end{equation}
Substituting this expectation into the marginal vector field yields the score-based representation of the flow:
\begin{equation}
    \mathbf{v}(\bz_t, t) = \frac{\bz_t}{t} + \frac{1-t}{t} \nabla_{\bz_t} \log p_t(\bz_t) 
\end{equation}
The term $\nabla_{\bz_t} \log p_t(\bz_t)$ ensures the trajectory is continuously pulled toward high-probability regions, allowing $p_{\text{data}}(\bz^{(k+1)})$ to approach $1$ as $t \to 1$.

However, applying the unguided field $\mathbf{v}$ could discard the optimization progress of Phase 1.
By substituting the score-based field into our guided formulation (e.g., using Option B from Eq.~\ref{eq:grad_guidance}), the effective integration dynamic becomes:
\begin{equation}
    \tilde{\mathbf{v}}_\theta(\mathbf{z}_t, t, c) \approx \underbrace{ \frac{\bz_t}{t} + \frac{1-t}{t} \nabla_{\bz_t} \log p_t(\bz_t) }_{\text{Manifold Correction}} - \underbrace{ \lambda \nabla_{\bz_t} \left( \mathcal{O}\big(\mathcal{D}(\bz_t)\big) - c \right)^2 }_{\text{Optimization Preservation}} 
\end{equation}
The score-based term acts as a correcting force that maximizes $p_{\text{data}}$ to correct the manifold drift, while the guidance term acts as a soft constraint, attracting this correction to the objective $c$ attained in Phase 1.

\subsection{Ablation study: guidance during flow matching}
\label{sec:appendix_ablation_guidance}

We evaluate the necessity of the guidance during flow-matching on the chair volume minimization task (3DShape2VecSet) targeting a 70\% reduction.
We compare our guided vector field $\tilde{\mathbf{v}}_\theta$ against an unguided baseline $\mathbf{v}_\theta$ during the Phase 2 correction (Equation \ref{eq:forward_noising}).


\begin{table}[htbp]
    \centering
    \caption{Ablation study on the effect of guidance in the flow-based correction (Phase 2). Evaluated on the chair volume optimization task (3DShape2VecSet) targeting a 70\% volume reduction.}
    \label{tab:ablation_guidance}
    \vspace{2mm}
    \resizebox{\linewidth}{!}{
    \begin{tabular}{lcc}
        \toprule
        \textbf{Method} & \textbf{Epochs to reach 70\% Vol. Red.} & \textbf{Success Rate (after 1000 epochs)} \\
        \midrule
        Ours (without guidance) & 737.57 & 70\% \\
        \textbf{Ours (with guidance)} & \textbf{335.18} & \textbf{100\%} \\
        \bottomrule
    \end{tabular}
    }
\end{table}

As shown in Tab.~\ref{tab:ablation_guidance}, the unguided field $\mathbf{v}_\theta$ blindly pulls the latent code toward the unconditional data manifold, washing out the progress achieved during the gradient descent phase.
Consequently, it requires 737.57 epochs on average and fails to reach the target for 30\% of shapes within 1000 epochs.
By contrast, our guided flow $\tilde{\mathbf{v}}_\theta$ prevents this wash-out, converging over three times faster (335.18 epochs) with a 100\% success rate.

\subsection{Chair volume optimization}
\label{app:chair}

In this section, we present a more detailed explanation of our experimental setup.
Furthermore, we provide extended results to further demonstrate the scalability and effectiveness of our method.

\subsubsection{Implementation details}
\label{app:chair_setup}
In this section, we provide comprehensive details regarding the dataset preprocessing, model architectures, and the exact optimization setup used for the chair volume minimization experiment described in the main text.

\paragraph{Dataset and Preprocessing.}
We utilize the chair category from the ShapeNet dataset \cite{Chang15}.
All meshes are preprocessed to be strictly watertight, canonically aligned, and normalized to fit within a bounding box of $[-1, 1]^3$.
We use a split of 2,419 chair shapes for training our shape priors and Flow Matching models.

To evaluate our optimization framework across different levels of latent complexity, we utilize two distinct generative priors and train a specific Flow Matching model for each.

\textbf{DeepSDF Architecture Details.} 
For a simple, continuous latent representation, we utilize the DeepSDF architecture \cite{Park19c}. 
Chair shapes are encoded into a single latent vector of dimension $D = 256$. 
The decoder is a multi-layer perceptron consisting of 8 layers with a hidden dimension of 512. 
This prior is trained for 2000 epochs using a batch size of 16 and the Adam optimizer, following a multi-step learning rate schedule.
The corresponding Flow Matching model operating on this vector space is parameterized by a Residual MLP. 
Crucially, this model is conditioned on the target shape volume using Classifier-Free Guidance (CFG) \cite{Ho22a}.
It is trained for 5000 epochs with a batch size of 256 and the Adam optimizer with a constant learning rate of 1e-4.

\textbf{3DShape2VecSet Architecture Details.} 
For a more complex, structured representation, we follow the 3DShape2VecSet architecture \cite{Zhang23d}.
The generative prior is configured with 12 layers, a hidden dimension of 512, and 8 attention heads of dimension 64. 
Here, chair shapes are encoded into a latent space represented by a vector set of $N = 256$ vectors, each with a dimensionality of $D = 8$. 
While the base architecture originally utilizes 512 vectors, we downscaled this set; since our shape prior encodes only a single shape category.
The prior is trained for 2000 epochs using a batch size of 32, the AdamW optimizer with a learning rate of 5e-5, and a Warmup Cosine schedule.
Its corresponding Flow Matching model operates directly on the latent set and is parameterized by a Transformer-based architecture \cite{Zhang23d}. 
Similar to the DeepSDF baseline, it is conditioned on the target volume via CFG.
This model is trained for 2000 epochs with a batch size of 32 and the AdamW optimizer with a learning rate of 1e-4, also utilizing a Warmup Cosine schedule.

\paragraph{Differentiable Volume Estimation.}
\label{app:volume_est}
The volume of a shape can be directly approximated using its Signed Distance Function (SDF).
To optimize the shape's volume while allowing gradients to flow back to the latent representation, we formulate a differentiable volume approximation.
During the optimization loop, we query the generative prior to predict the SDF values on a uniform 3D grid $\mathcal{G}$ of resolution $64^3$ spanning the canonical bounding box $[-1, 1]^3$.
To approximate the binary inside/outside occupancy in a smooth, differentiable manner, we apply a sharp sigmoid function to the negative SDF values.
The estimated volume $V$ given a latent code $\mathbf{z}$ is computed as the mean occupancy across all grid points:

\begin{equation}
V(\mathbf{z}) = \frac{1}{|\mathcal{G}|} \sum_{\mathbf{x} \in \mathcal{G}} \sigma\big(-\alpha \cdot f_\theta(\mathbf{x}, \mathbf{z})\big)
\end{equation}
where $f_\theta(\mathbf{x}, \mathbf{z})$ is the predicted SDF value at point $\mathbf{x}$, $\sigma$ is the standard sigmoid function, and $\alpha = 100$ is a sharpness factor that scales the SDF to approximate a step function at the surface boundary.

\paragraph{Optimization Hyperparameters}
The complete set of hyperparameters used in Algorithm~\ref{alg:algo} for this volume minimization experiment, comparing the DeepSDF and 3DShape2VecSet latent representations, is summarized in Tab.~\ref{tab:chair_optim_hyperparams}.

\begin{table}[htbp]
    \centering
    \caption{
    Optimization hyperparameters for chair volume minimization across different latent representations.
    }
    \label{tab:chair_optim_hyperparams}
    \vspace{2mm}
    \small
    \setlength{\tabcolsep}{8pt}
    \begin{tabular}{l | l c c}
        \toprule
        \textbf{Phase} & \textbf{Parameter} & \textbf{DeepSDF} & \textbf{3DShape2VecSet} \\ \midrule
        \multirow{2}{*}{Gradient Descent} & Number of steps ($M$) & 30 & 15 \\
        & Learning rate ($\eta$) & 0.001 & 0.005 \\ \midrule
        \multirow{5}{*}{Flow Correction} & Noise level ($\tau$) & 0.1 & 0.2 \\
        & Flow Matching start time ($t_{\text{start}}$) & 0.9 & 0.8 \\
        & Total ODE schedule steps & 100 & 100 \\
        & Actual ODE steps ($N$) & 10 & 10 \\
        & CFG scale ($s$) & 1.0 & 1.0 \\
        \bottomrule
    \end{tabular} 
\end{table}

\newpage
\subsubsection{Additional results}
\label{app:chair_results}

\begin{figure}[H]
  \centering
  \includegraphics[width=0.8\textwidth]{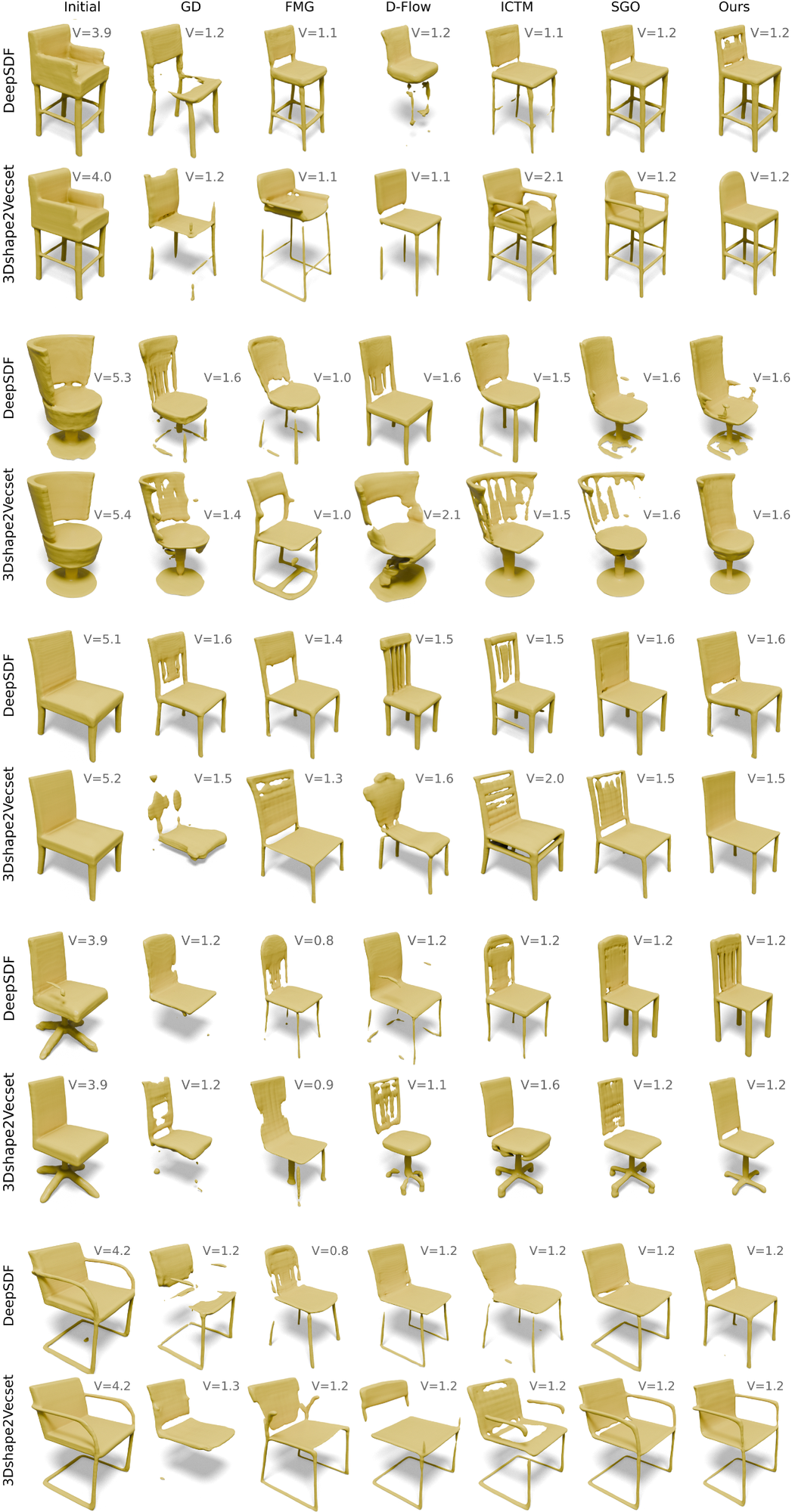}
\caption{Additional qualitative results for chair volume optimization.
We compare the original shape (Initial) with different baseline methods and our approach, using both DeepSDF and 3DShape2VecSet.
In the figure, $V$ indicates the measured volume of the shape.}
\label{fig:app_chair_shape}
\end{figure}

\paragraph{Additional Qualitative Results.}
In Figure \ref{fig:app_chair_shape}, we show more visual examples for the chair volume optimization.
As discussed in the main text, these examples show how baseline methods often break the chair's structure to reach the volume target (for example, by creating floating parts or removing legs), especially when using the more complex 3DShape2VecSet model.
For the exact same optimization target, our method successfully preserves the shapes without breaking them.
Furthermore, our approach produces higher quality results on 3DShape2VecSet because our framework manages to use the model's full expressivity without suffering from severe manifold drift.
This proves that complex models are very powerful, as long as the manifold drift problem is controlled.

\begin{figure}[h]
    \centering
    \includegraphics[width=\linewidth]{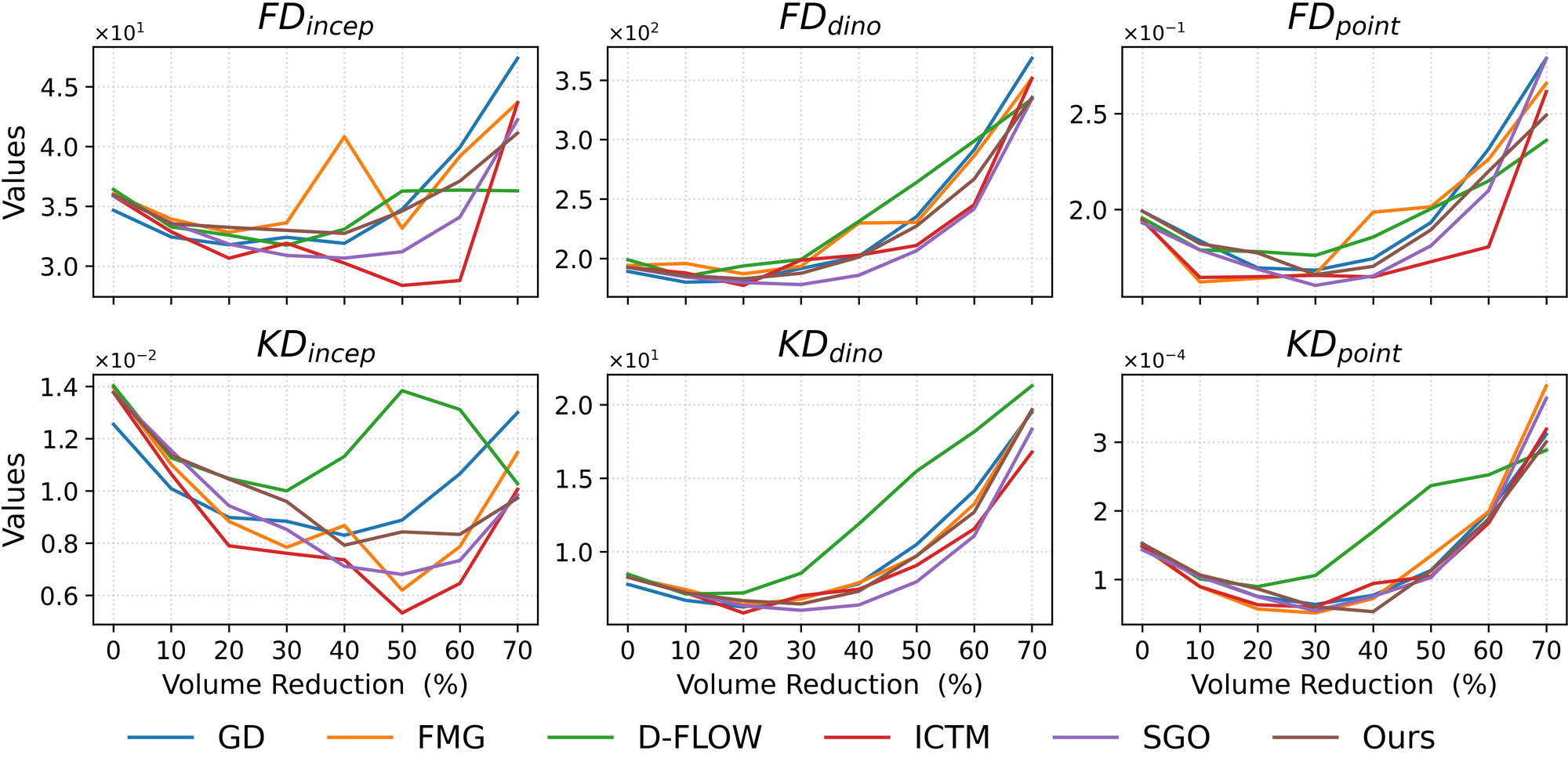}
    \caption{Evolution of generative metrics (FD and KD) for the \textbf{chair} optimization with  \textbf{DeepSDF} across different volume reduction targets.}
    \label{fig:app_chair_deepsdf_metrics_evolution}
\end{figure}

\begin{figure}[H]
    \centering
    \includegraphics[width=\linewidth]{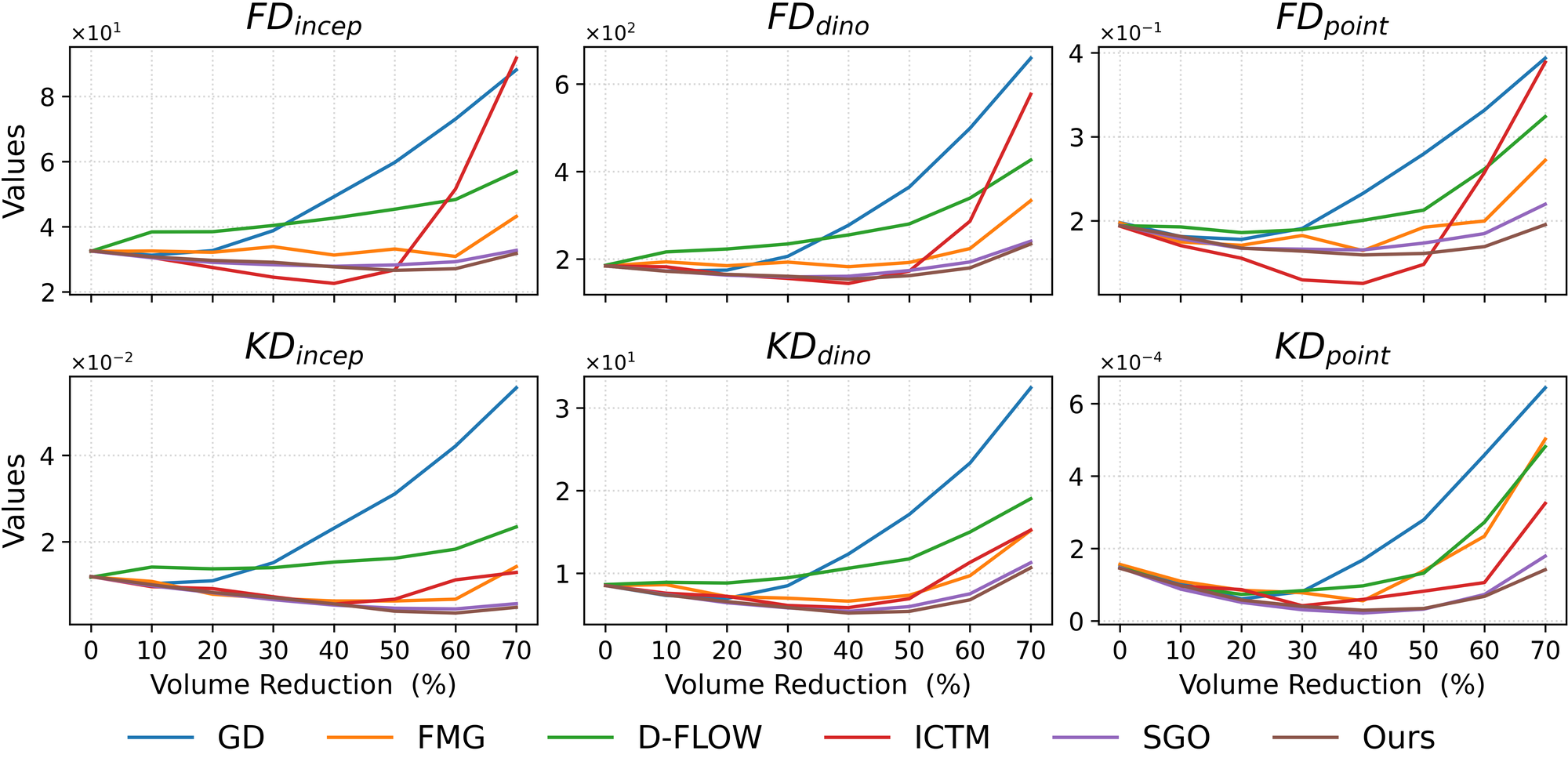}
    \caption{Evolution of metrics (FD and KD) for the \textbf{chair} optimization with  \textbf{Vecset} across different volume reduction targets.}
    \label{fig:app_chair_vecset_metrics_evolution}
\end{figure}

\paragraph{Generative Quality vs. Optimization Targets.}
Figure \ref{fig:app_chair_deepsdf_metrics_evolution} and \ref{fig:app_chair_vecset_metrics_evolution} show evolution of metrics (FD and KD) for the chair optimization respectively with DeepSDF and 3DShape2VecSet across different volume reduction targets.

On Figure \ref{fig:app_chair_deepsdf_metrics_evolution}, we can see that when using a simpler model like DeepSDF, all methods exhibit similar behavior.
In this simpler latent space, manifold drift is not a significant issue, and the metrics remain relatively stable across different volume reduction targets.

Conversely, transitioning to a more complex generative model such as 3DShape2VecSet reveals a stark difference, as shown in Figure \ref{fig:app_chair_vecset_metrics_evolution}.
As the volume reduction objective becomes more aggressive, the metrics for most baseline methods increase sharply, highlighting the increased difficulty of optimization and a severe manifold drift in highly expressive models.
As discussed in the experiments section, SGO also proves to be quite robust in this specific setting, maintaining relatively low metric values. However, this robustness does not generalize to our other experiments involving more complex optimization objectives.
Ultimately, our method consistently exhibits the most controlled increase overall, demonstrating its robust effectiveness in mitigating manifold drift and preserving shape quality across different tasks and within highly complex latent spaces.

\newpage
\subsection{Car drag optimization}
\label{app:car}

In this section, we present a more detailed explanation of our experimental setup.
Furthermore, we provide extended results to further demonstrate the scalability and effectiveness of our method.

\subsubsection{Implementation details}
\label{app:car_setup}
In this section, we provide comprehensive details regarding the dataset preprocessing, model architectures, and the surrogate model used for the aerodynamic drag reduction experiment described in the main text (Section \ref{sec:car}).

\paragraph{Dataset and Preprocessing.}
We utilize the car category from the ShapeNet dataset \cite{Chang15}.
We preprocess the meshes to be strictly watertight and simulate the resulting dataset of 1036 cars using OpenFOAM to compute the drag.
To ensure consistent aerodynamic evaluation, all meshes are aligned and normalized into a bounding box of $[-1, 1]^3$.
Finally, we randomly split this dataset into 836 shapes for training the shape prior and 200 shapes for the test set.

\paragraph{Architecture Details.}
For our generative prior, we follow the 3DShape2VecSet architecture \cite{Zhang23d}, though we adapt its dimensions to better suit our single-category training.
The model is configured with 12 layers, a hidden dimension of 512, and 8 attention heads of dimension 64.
Car shapes are encoded into a latent space represented by a vector set of $N = 256$ vectors, each with a dimensionality of $D = 8$.
While the base architecture originally utilizes 512 vectors, we downscaled this set; since our shape prior encodes only a single shape category.
The prior is trained for 2000 epochs using a batch size of 16, the AdamW optimizer with a learning rate of 5e-5, and a Warmup Cosine schedule.

Our Flow Matching model operates directly on this latent set and is parametrized by a Transformer-based architecture following the 3DShape2VecSet Denoising Network architecture \cite{Zhang23d}.
The Flow Matching model is trained with a drag coefficient ($C_d$) conditioning using Classifier-Free Guidance (CFG) \cite{Ho22a}.
The model is trained for 2000 epochs with a batch size of 32 and the AdamW optimizer with a learning rate of 1e-4, also following a Warmup Cosine schedule.

\paragraph{Surrogate Model}
Computing the actual aerodynamic drag coefficient ($C_d$) during the optimization loop is computationally intractable and non-differentiable via standard CFD solvers.
To guide our shape optimization, we train a surrogate model to provide a differentiable estimate of the surface pressure $\hat{p}(\mathbf{x})$ and the resulting drag. 

The drag force is computed as the surface integral of the air pressure:
\begin{equation}
\text{drag}_p(\mathcal{S}) = \iint_{\mathcal{S}} -n_x(\mathbf{x}) \cdot p(\mathbf{x}) \, d\mathcal{S}(\mathbf{x})
\end{equation}
where $p(\mathbf{x})$ is the pressure and $\mathbf{n}(\mathbf{x})$ is the surface normal at point $\mathbf{x}$, with $n_x$ being its component along the $x$-axis (directed from the front to the back of the car).
This value is then normalized to obtain the drag coefficient $C_d$ \cite{Munson13}. Following standard practice for cars, we ignore friction drag as it remains negligible compared to pressure drag.

Our surrogate model is a Graph Neural Network \cite{Baque18} utilizing GraphSage convolution layers \cite{Hamilton17, Bonnet22}.
It is trained to predict $\hat{p}(\mathbf{x})$ directly on the mesh vertices. 
To generate the ground-truth training labels, we simulate the airflow around the dataset of cars using the OpenFOAM (OpenCFD) solver \cite{OpenFoam}, following the well-established DrivAer model setup \cite{Wieser14}. 

The optimization objective is then defined by minimizing the drag coefficient predicted through the surrogate model:
\begin{equation}
\mathcal{L}_{\text{drag}} = C_d(\text{MC}(f_\theta, \mathbf{z}))
\end{equation}
where $\text{MC}$ denotes the Marching Cubes algorithm \cite{Lewiner03}, $f_\theta$ is the decoder, and $\mathbf{z}$ represents the latent representation.
This setup allows gradients to be backpropagated from the drag loss through the meshing step \cite{Remelli20b} and the generative prior to update the latent representation.

\paragraph{Optimization Hyperparameters}
Tab.~\ref{tab:car_optim_hyperparams} details the complete set of hyperparameters used in Algorithm~\ref{alg:algo}.
These parameters govern the alternating scheme between gradient-based drag minimization and generative flow correction.


\begin{table}[htbp]
    \centering
    \caption{
    Optimization hyperparameters for aerodynamic drag reduction with our framework.
    }
    \label{tab:car_optim_hyperparams}
    \vspace{2mm}
    \small
    \setlength{\tabcolsep}{10pt}
    \begin{tabular}{l | l c}
        \toprule
        \textbf{Phase} & \textbf{Parameter} & \textbf{Value} \\ \midrule
        \multirow{2}{*}{Gradient Descent} & Number of steps ($M$) & 20 \\
        & Learning rate ($\eta$) & 0.01 \\ \midrule
        \multirow{5}{*}{Flow Correction} & Noise level ($\tau$) & 0.2 \\
        & Flow Matching start time ($t_{\text{start}}$) & 0.8 \\
        & Total ODE schedule steps & 100 \\
        & Actual ODE steps ($N$) & 20 \\
        & CFG scale ($s$) & 1.0 \\
        \bottomrule
    \end{tabular} 
\end{table}

\subsubsection{Validation of true aerodynamic performance}
\label{app:simu}

To validate the physical realism of the car drag reduction experiment, we conducted Computational Fluid Dynamics (CFD) simulations using OpenFOAM \cite{OpenFoam}.
For this evaluation, we simulated 20 car shapes per method, with each shape optimized to target a 30\% drag reduction according to the surrogate model. 


\begin{table}[htbp]
    \centering
    \caption{
        OpenFOAM simulation results at 30\% drag reduction (20 cars simulate per method).}
    \label{tab:car_simulation_result}
    \vspace{2mm}
    \small
    \setlength{\tabcolsep}{6pt} 
    \resizebox{0.9\textwidth}{!}{
    \begin{tabular}{l | c c c c c c}
        \toprule
        \textbf{Metric} & \textbf{GD} & \textbf{FMG} & \textbf{D-Flow} & \textbf{ICTM} & \textbf{SGO} & \textbf{Ours} \\ \midrule
        \textbf{True Reduction (\%)} & 19.38 & 20.63 & 14.47 & 20.98 & 19.27 & 17.35 \\
        \textbf{Surrogate Error (MAE)} $\times 10^{2}$ & 4.8 & 5.2 & 6.1 & 4.9 & 4.5 & 4.1 \\
        \bottomrule
    \end{tabular} 
    }
\end{table}

\begin{figure}[h]
  \centering
  \vspace{-1mm} 
  \includegraphics[width=\textwidth]{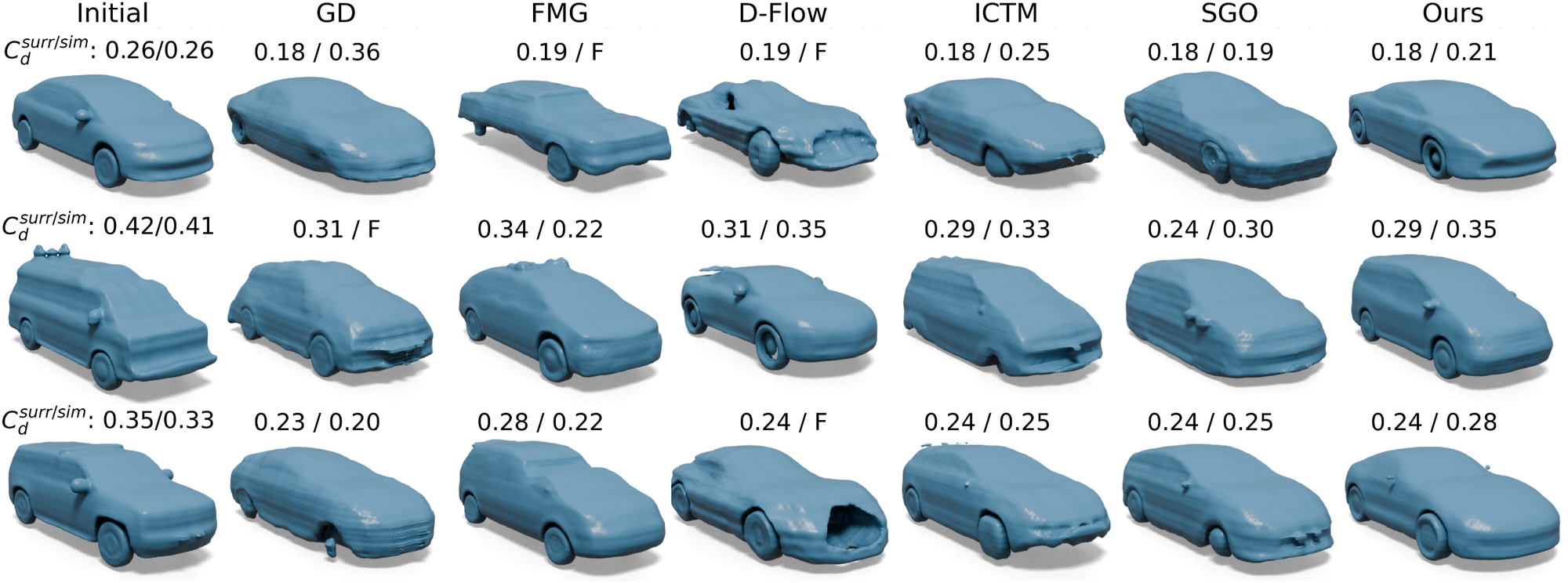} 
\caption{
Comparison of optimized car shapes. For each shape, we provide the drag coefficient ($C_d$) evaluated by the surrogate model and the actual OpenFOAM simulation (surr / sim). 'F' indicates that the OpenFOAM simulation failed, typically due to severe, non-physical geometric distortions preventing proper meshing.
}
\label{app:car_shape_drag_simu}
\end{figure}

The quantitative results are summarized in Tab.~\ref{tab:car_simulation_result}.
Our method achieves a true physical drag reduction of 17.35\%.
This gap between the targeted and simulated reduction is expected for two main reasons.
First, surrogate models are inherently imperfect approximations, which is standard in this domain.
Second, our surrogate model is trained to predict only the pressure component of the drag, whereas the OpenFOAM simulation computes the total drag, which encompasses both pressure and skin friction components.

Despite this expected systematic shift, we observe that certain baselines, such as FMG (20.63\%) or SGO (19.27\%), exhibit a slightly higher true drag reduction (roughly 3\% to 4\% better than ours).
However, this numerical advantage comes at the complete cost of shape validity. As illustrated in Fig.~\ref{app:car_shape_drag_simu}, these baselines achieve a lower physical drag by fundamentally destroying the vehicle's geometry rather than optimizing a realistic car.
They devolve into over-smoothed blobs, absorb wheels into the chassis, or introduce severe topological artifacts. While such degenerate, featureless shapes naturally offer less aerodynamic resistance in physics, they no longer represent valid or functional vehicles.
In contrast, our method preserves high-quality, realistic car structures while still achieving substantial aerodynamic improvements.

Furthermore, Tab.~\ref{tab:car_simulation_result} indicates that our approach maintains the lowest surrogate prediction error (MAE of $4.1 \times 10^{-2}$). We hypothesize that this better alignment between the surrogate predictions and the actual CFD simulations stems from our flow-guided approach. Navigating the optimization closer to the valid data manifold likely keeps the generated shapes in regions where the surrogate remains accurate, whereas manifold drift in baseline methods can lead to out-of-distribution shapes where the surrogate's reliability decreases.

\subsubsection{Additional results}
\label{app:car_results}

\begin{figure}[H]
  \centering
  \includegraphics[width=\textwidth]{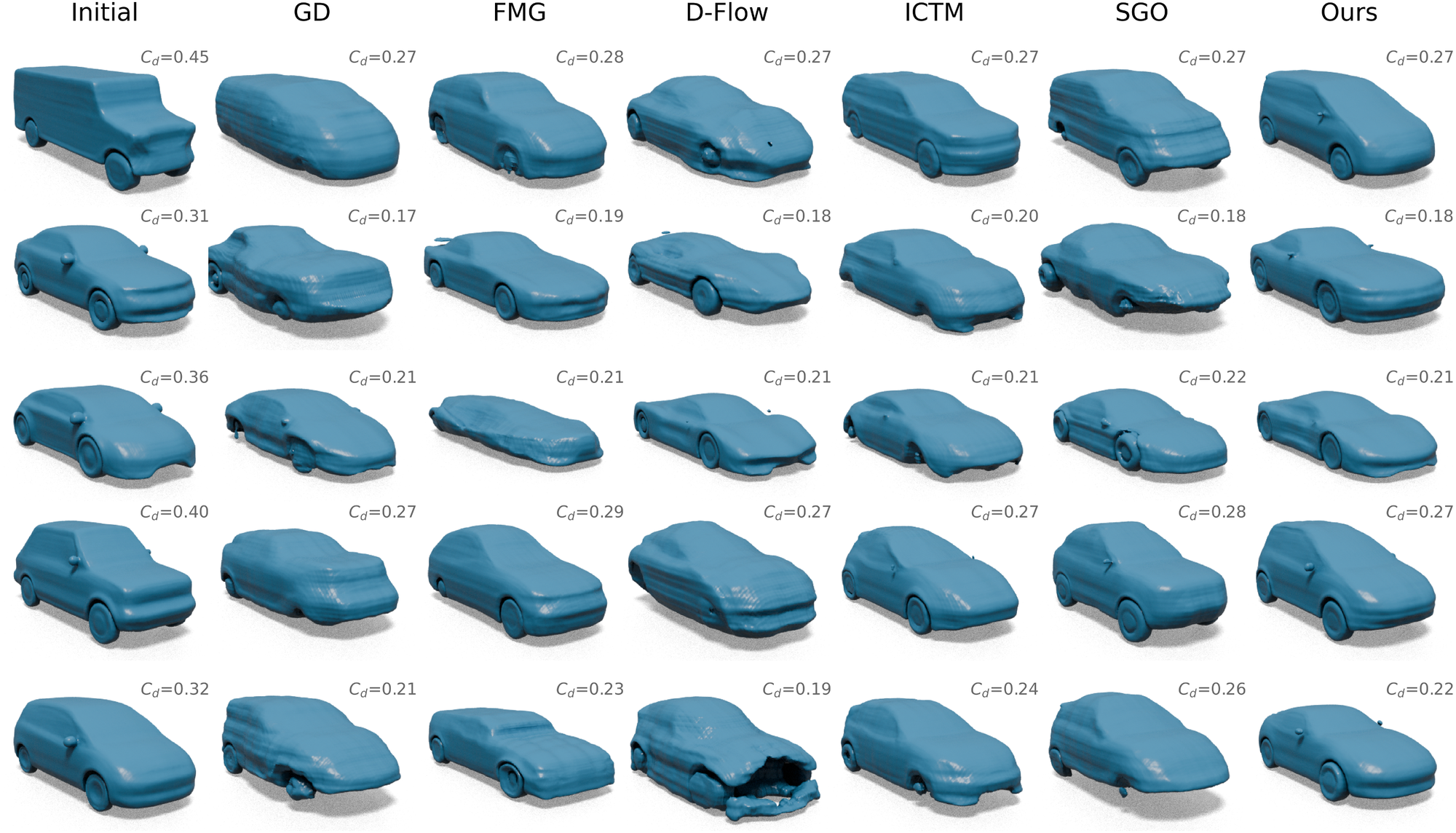} 
\caption{
Additional qualitative comparisons of car shapes optimized for a targeted 30\% drag reduction.
We report the surrogate drag coefficient ($C_d$) for the initial shapes and the outputs of each optimization method.
}
\label{fig:app_car_shape}
\end{figure}

\paragraph{Additional Qualitative Results.}
We provide additional qualitative comparisons of optimized car shapes in Fig.~\ref{fig:app_car_shape}, where each method was tasked with achieving a targeted 30\% drag reduction according to the surrogate model.
As observed, while all methods successfully reduce the surrogate drag coefficient ($C_d$) to reach this target margin, the numerical performance of the baselines often comes at the severe cost of geometric integrity.
Baseline methods frequently suffer from manifold drift, producing degenerate shapes such as over-smoothed bodies, absorbed wheels, or non-physical topological artifacts (e.g., see GD and D-Flow).
In contrast, our approach consistently maintains the structural validity and visual realism of the vehicles while achieving the targeted aerodynamic efficiency.

\begin{figure}[H]
    \centering
    \includegraphics[width=\linewidth]{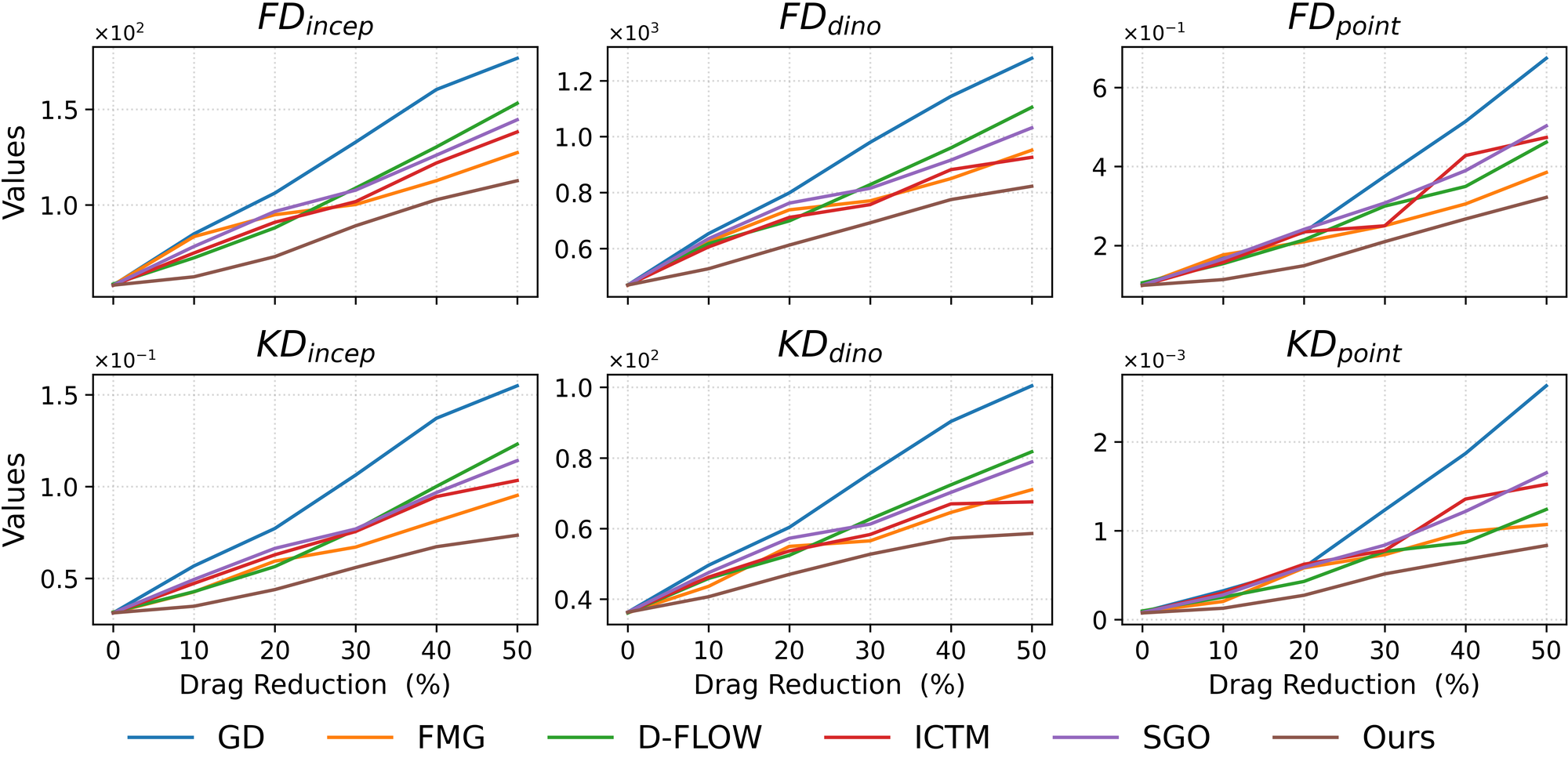}
    \caption{Evolution of metrics (FD and KD) across different drag reduction targets.}
    \label{fig:app_car_metrics_evolution}
\end{figure}

\paragraph{Generative Quality vs. Optimization Targets.}
To further analyze the trade-off between aerodynamic drag reduction and shape quality, we evaluate the evolution of Fréchet Distance (FD) and Kernel Inception Distance (KD) across increasingly strict drag reduction targets.
As illustrated in Fig.~\ref{fig:app_car_metrics_evolution}, pushing the optimization to extreme high reductions naturally degrades the generative metrics for all methods, reflecting a deviation from the reference shape distribution.
However, our method consistently exhibits a significantly slower degradation, maintaining much lower FD and KD scores compared to the baselines (GD, FMG, D-Flow, ICTM, and SGO) across all evaluated feature extractors (Inception, DINOv2, and PointBert).
This quantitative analysis confirms that our framework effectively attracts the optimization trajectory to the manifold of plausible car shapes.

\subsection{Compliance optimization in Hunyuan3D}
\label{app:Hunyuan3D}

In this section, we present a more detailed explanation of our experimental setup.
Furthermore, we provide extended results to further show the effectiveness of our method.

\subsubsection{Implementation details}
\label{app:Hunyuan3D_setup}

\paragraph{Foundation Model Prior: Hunyuan3D.}
To demonstrate the scalability of our approach, we use Hunyuan3D, a state-of-the-art generative foundation model capable of synthesizing high-fidelity 3D assets.
In our pipeline, the initial 3D shapes are instantiated using the model's image-to-3D capabilities, generating complex geometries conditioned on 2D input images.
Unlike category-specific models, Hunyuan3D captures immense structural diversity by representing shapes as continuous $4096 \times 64$ latent codes.
While this vast, high-dimensional latent space is natively designed for feed-forward generation, our method successfully repurposes it as a highly expressive optimization domain.

\paragraph{Compliance Formulation}
\label{app:compliance}
To reinforce the generated objects against external forces, we adopt the differentiable physics optimization framework introduced by PhysiOpt \cite{Zhan25a}.
Assuming static equilibrium, the object's displacement $u$ is governed by the linear system $K(\pi)u(\pi) = f$, where $f$ represents the external nodal forces and $K(\pi)$ is the stiffness matrix parameterized by the latent variables $\pi$.
The compliance $C$ is calculated as the work done by the external forces, defined as $C = f^T u$.
To differentiably compute the compliance without explicit mesh extraction, PhysiOpt \cite{Zhan25a} maps the shape's implicit field onto a sparse voxel grid.
Each finite element's continuous density $\rho_e$ is derived from the implicit field at its nodes and scales its local stiffness matrix via $K_e = \rho_e K_e^{solid}$.
Assembling these yields the global stiffness matrix $K$, enabling the finite element method (FEM) to compute displacements $u$ from the static equilibrium $Ku = f$ and directly evaluate the shape's compliance.
This voxel-based formulation enables seamless, fast gradient-based updates throughout the FEM solve, allowing us to backpropagate the physical loss directly into Hunyuan3D's high-dimensional latent space.

\paragraph{Simulation Setup.}
We use a finite element voxel resolution of $r=64$ to accurately resolve the intricate structural details synthesized by Hunyuan3D.
For the mechanical simulation, we assume a homogeneous isotropic material typical of rigid plastics or wood, setting the Young's modulus to $E = 0.5$ GPa and the Poisson's ratio to $\nu = 0.3$.
Finally, to replicate a natural resting state, the boundary conditions are defined by completely anchoring the bottom 5\% of the object's vertical bounding box (Y-axis) to simulate contact with the ground, while external loads are applied to category-specific functional regions.

\paragraph{Quantitative Evaluation on the ABO and BIKED Datasets.}
To move beyond isolated qualitative examples and rigorously evaluate the generative quality of our optimized shapes, we established a quantitative benchmark using the ABO dataset \cite{Collinsand21a} and the BIKED dataset \cite{Regenwetter22}.
We specifically extracted real-world images of chairs and tables from ABO, and bicycles from BIKED, filtering the data to retain only valid and high-quality images.
These images were then used to condition Hunyuan3D to generate the initial 3D shapes.
To automate the physical optimization across this diverse set of objects, we defined generic forces, systematically targeting the functional surfaces of the objects (i.e., the seat for chairs, the top surface for tables, and the saddle for bicycles).
In total, we optimized 100 objects for chairs and tables, and bicycles, across different baseline methods, evaluating them at fixed compliance reduction thresholds.
To compute our generative metrics (FID and KID), we compared these optimized outputs against reference sets of unoptimized generated shapes for each category (including 162 chairs, 159 tables, 183 bicycles).

\paragraph{Optimization Hyperparameters}
Tab.~\ref{tab:physiopt_optim_hyperparams} details the complete set of hyperparameters used in Algorithm~\ref{alg:algo}.
These parameters govern the alternating scheme between gradient-based compliance minimization and generative flow correction.


\begin{table}[htbp]
    \centering
    \caption{
    Optimization hyperparameters for PhysiOpt optimization on Hunyuan3D with our framework.
    }
    \label{tab:physiopt_optim_hyperparams}
    \vspace{2mm}
    \small
    \setlength{\tabcolsep}{10pt}
    \begin{tabular}{l | l c}
        \toprule
        \textbf{Phase} & \textbf{Parameter} & \textbf{Value} \\ \midrule
        \multirow{2}{*}{Gradient Descent} & Number of steps ($M$) & 5 \\
        & Learning rate ($\eta$) & 0.1 \\ \midrule
        \multirow{5}{*}{Flow Correction} & Noise level ($\tau$) & 0.4 \\
        & Flow Matching start time ($t_{\text{start}}$) & 0.6 \\
        & Total ODE schedule steps & 100 \\
        & Actual ODE steps ($N$) & 40 \\
        & Gradient Guidance Scale ($\lambda$) & 0.001 \\
        \bottomrule
    \end{tabular} 
\end{table}

\subsubsection{Additional results}
\label{app:Hunyuan3D_results}

\begin{figure}[H]
  \centering
  \includegraphics[width=\textwidth]{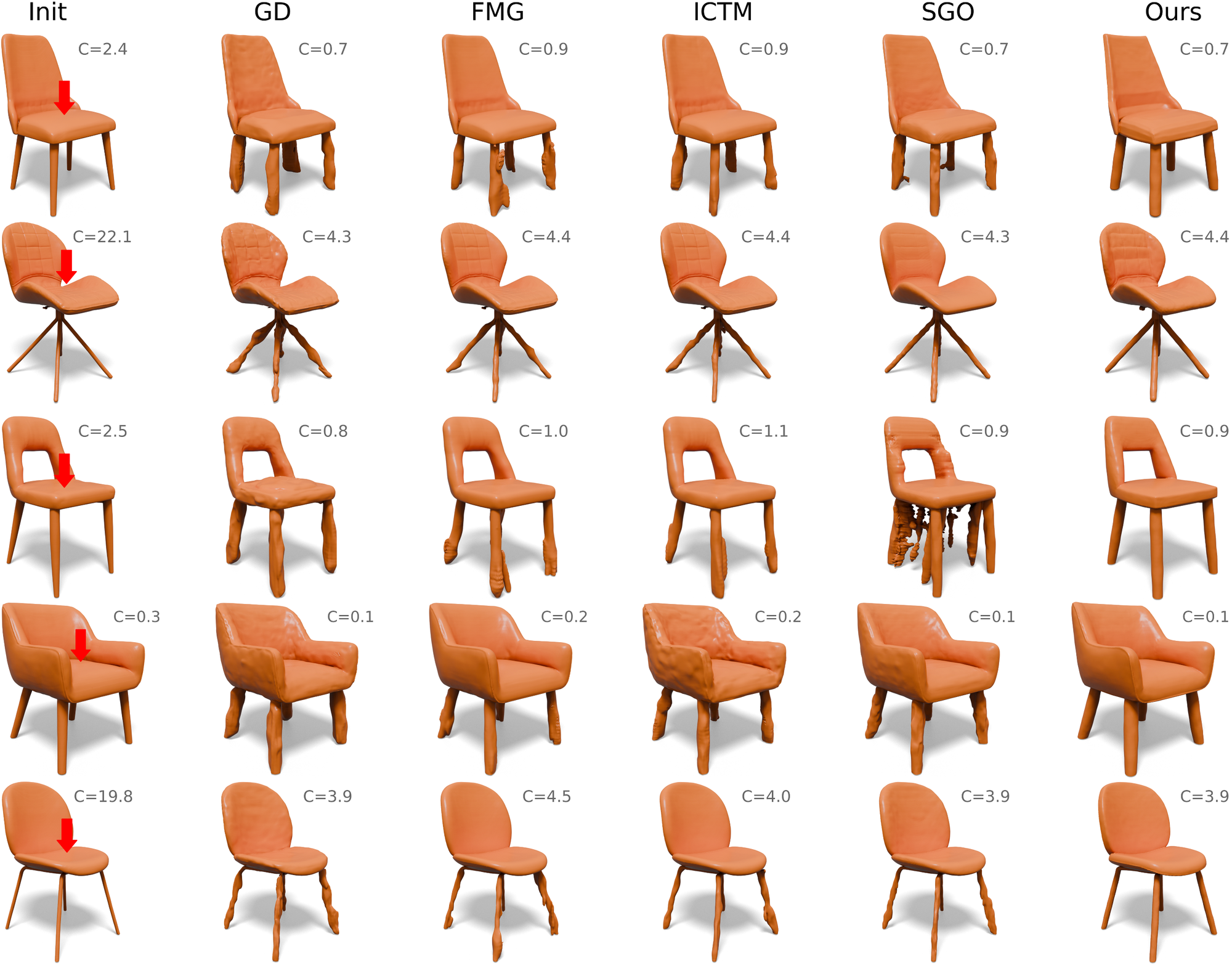} 
\caption{Qualitative result on the ABO dataset with chairs.
We show the initial shapes (Init) with the applied force (red arrow) and the optimized shapes produced by different methods (GD, FMG, ICTM, SGO, and Ours).
The compliance ($C$) is reported for each shape, where lower values indicate better structural stability.}
\label{fig:app_ago_chair_shape}
\end{figure}

\begin{figure}[H]
  \centering
  \includegraphics[width=\textwidth]{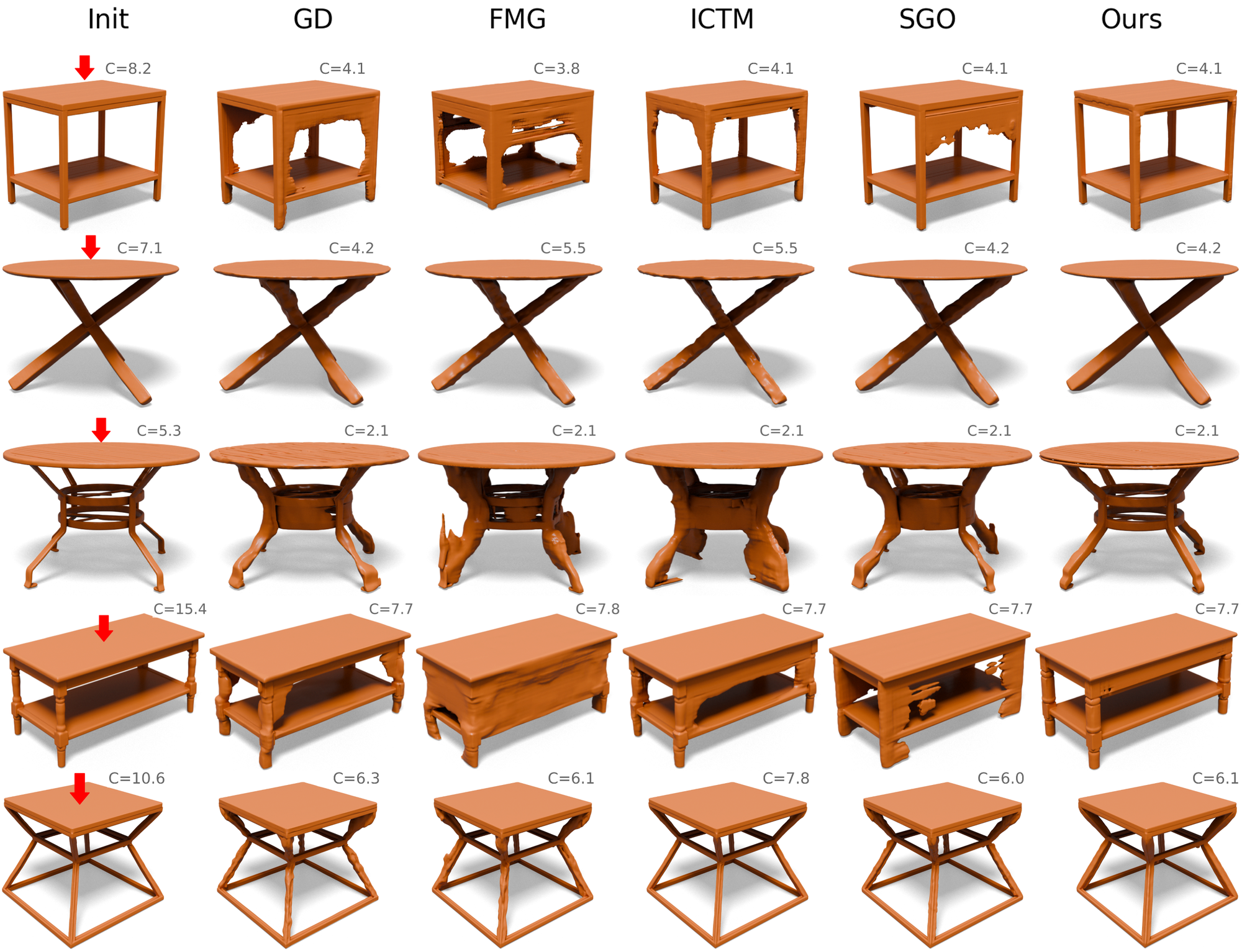} 
\caption{Qualitative result on the ABO dataset with tables.
We show the initial shapes (Init) with the applied force (red arrow) and the optimized shapes produced by different methods (GD, FMG, ICTM, SGO, and Ours).
The compliance ($C$) is reported for each shape, where lower values indicate better structural stability.}
\label{fig:app_ago_table_shape}
\end{figure}

\paragraph{ABO Dataset.}
To provide deeper insights into our quantitative evaluation, we present visual examples of the objects optimized during the experiments.
Fig.~\ref{fig:app_ago_chair_shape} and Fig.~\ref{fig:app_ago_table_shape} illustrate the structural modifications introduced by our optimization method across chairs and tables, respectively.
Our method achieves comparable compliance reduction to baselines while better preserving the original geometry.
In contrast, baseline approaches often introduce unnatural artifacts or bumpy surface distortions to minimize compliance.

In addition to the chairs reported in Tab.~\ref{tab:phyiopt_table_result}, we report in Tab.~\ref{tab:abo_table_result} the quantitative results for the ABO tables at 50\% compliance reduction.
Consistent with the other categories, our method outperforms all baselines across all metrics, while D-Flow runs out of memory.
It also best preserves the identity of the initial shapes, with an LPIPS of $0.04 \pm 0.03$, compared to $0.05 \pm 0.06$ for SGO, $0.08 \pm 0.08$ for ICTM, $0.12 \pm 0.13$ for GD, and $0.15 \pm 0.14$ for FMG.

\begin{table}[H]
    \centering
    \caption{
    Compliance ($C$) optimization on ABO tables at 50\% reduction.
    Scaling: $\text{KD}_{\text{incep}}$($\times10^3$), $\text{FD}_{\text{point}}$($\times 10^2$), $\text{KD}_{\text{point}}$($\times 10^5$).
    }
    \label{tab:abo_table_result}
    \vspace{2mm}
    \small
    \setlength{\tabcolsep}{4pt}
    \begin{tabular}{l | c c c c c c}
        \toprule
        \textbf{Method} & \textbf{FD}$_{\text{incep}}$ & \textbf{KD}$_{\text{incep}}$ & \textbf{FD}$_{\text{dino}}$ & \textbf{KD}$_{\text{dino}}$ & \textbf{FD}$_{\text{point}}$ & \textbf{KD}$_{\text{point}}$ \\ \midrule
        GD            & 81.52 & 19.31 & 684.09 & 32.19 & 51.98 & 31.80 \\
        FMG           & 69.18 & 13.71 & 530.29 & 18.93 & 47.96 & 17.40 \\
        D-Flow        & \multicolumn{6}{c}{\textit{OOM}} \\
        ICTM          & 67.11 & 9.43 & 506.05 & 15.56 & 45.87 & 9.50 \\
        SGO           & 55.56 & 4.83 & 402.84 & 12.43 & 37.36 & 1.80 \\
        \textbf{Ours} & \textbf{49.97} & \textbf{3.59} & \textbf{361.04} & \textbf{9.06} & \textbf{35.07} & \textbf{0.50} \\
        \bottomrule
    \end{tabular}
\end{table}

\begin{figure}[H]
    \centering
    \includegraphics[width=\linewidth]{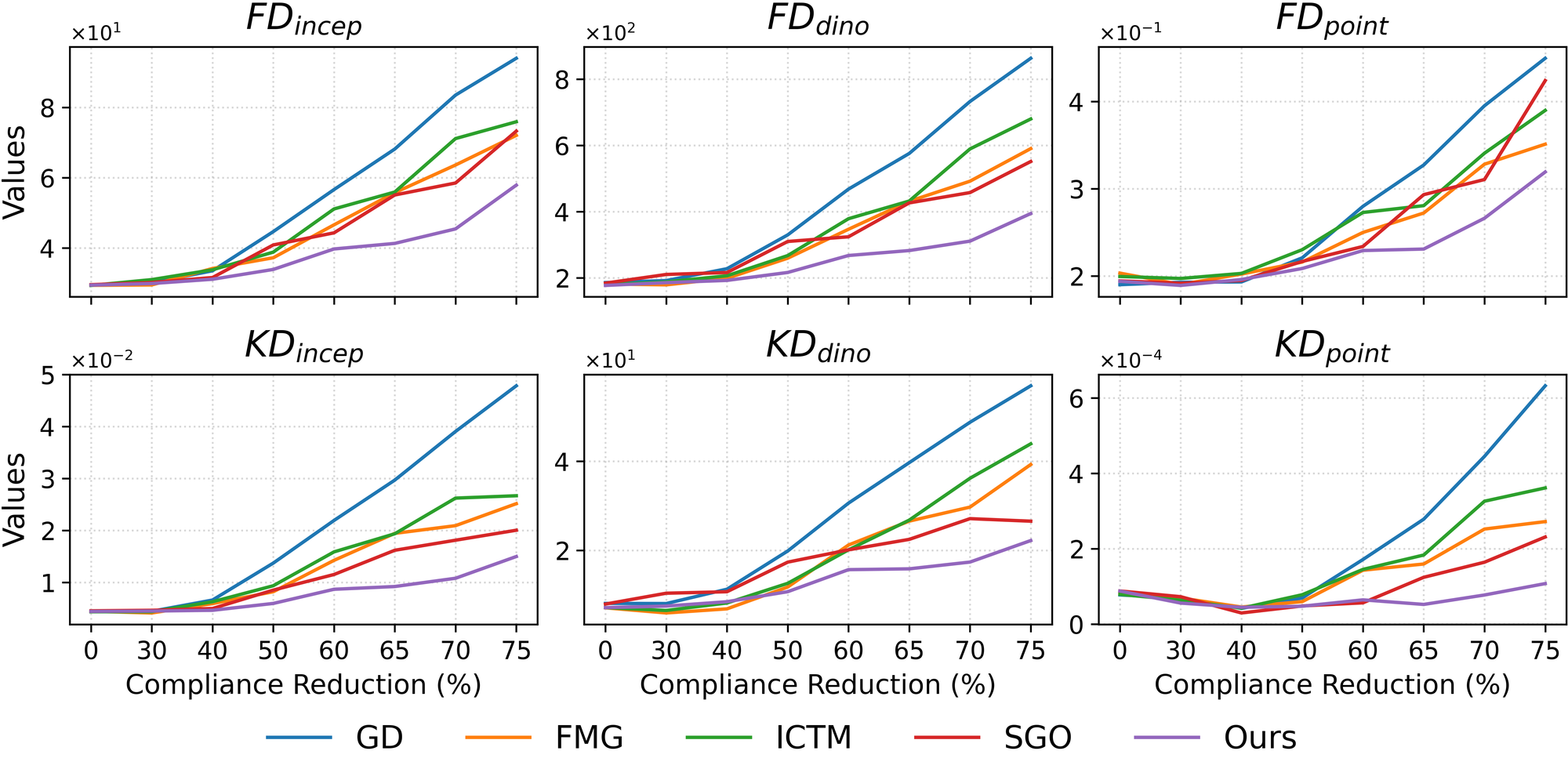}
    \caption{Evolution of generative metrics (FD and KD) for the \textbf{chair} category across different optimization thresholds.}
    \label{fig:app_chair_metrics_evolution}
\end{figure}

\begin{figure}[H]
    \centering
    \includegraphics[width=\linewidth]{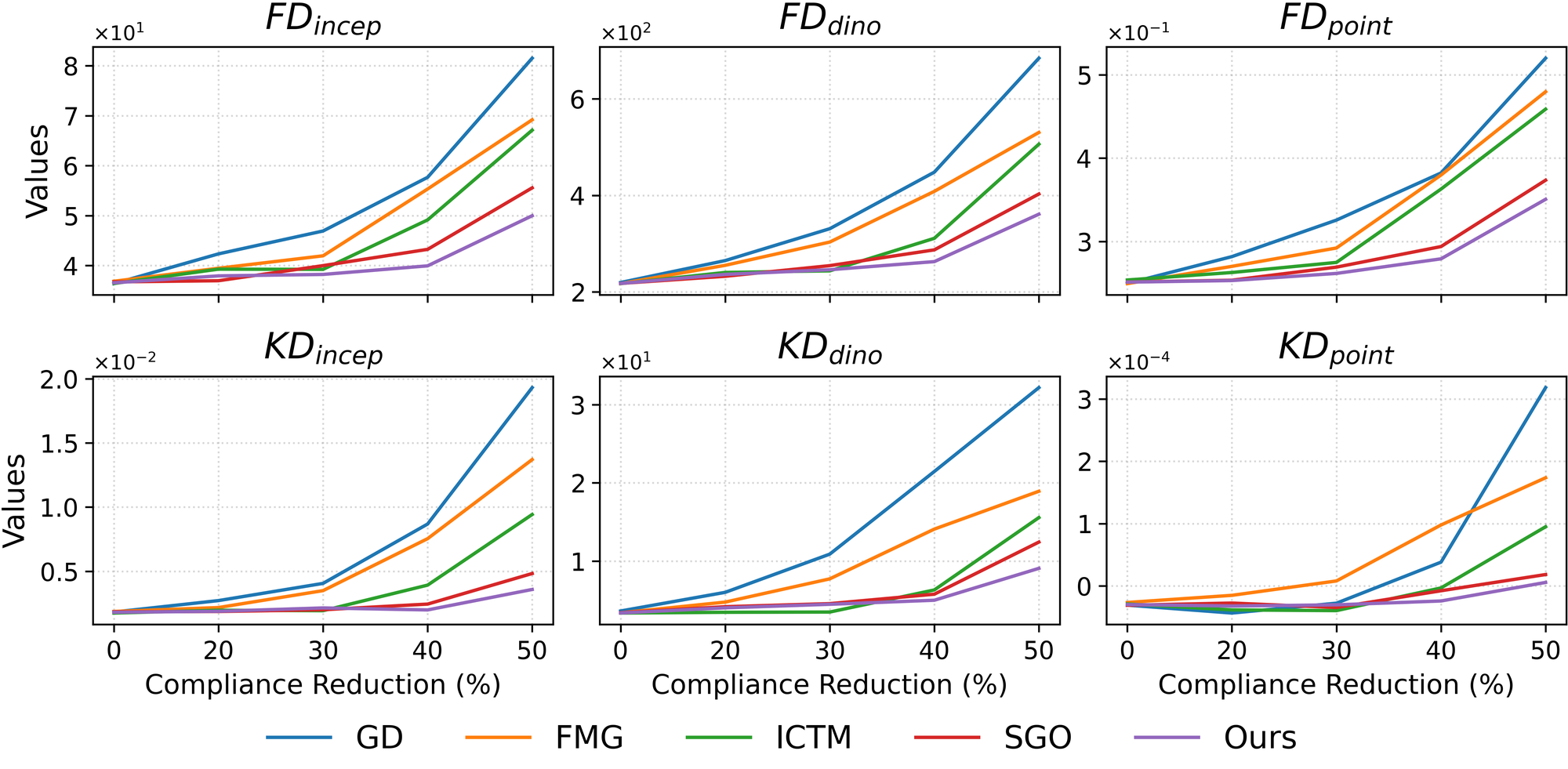}
    \caption{Evolution of generative metrics (FD and KD) for the \textbf{table} category across different compliance reduction targets.}
    \label{fig:app_table_metrics_evolution}
\end{figure}

\paragraph{Generative Quality vs. Optimization Targets.}
To further analyze the trade-off between physical reinforcement and shape quality, we evaluate the evolution of Fréchet Distance (FD) and Kernel Inception Distance (KD) across increasingly strict optimization thresholds.
We perform this analysis for both object categories, as illustrated in Fig.~\ref{fig:app_chair_metrics_evolution} for chairs and Fig.~\ref{fig:app_table_metrics_evolution} for tables.
As expected, pushing the optimization to extreme targets naturally degrades the generative metrics for all methods, reflecting a deviation from the reference shape distribution.
However, our method consistently exhibits a significantly slower degradation, maintaining much lower FD and KD scores compared to the baselines (GD, FMG, ICTM, and SGO) across all evaluated feature extractors (Inception, DINOv2, and PointBert).
This quantitative analysis confirms that our framework effectively attracts the optimization trajectory to a manifold of plausible shapes for both chairs and tables.

\begin{figure}[H]
  \centering
  \includegraphics[width=\textwidth]{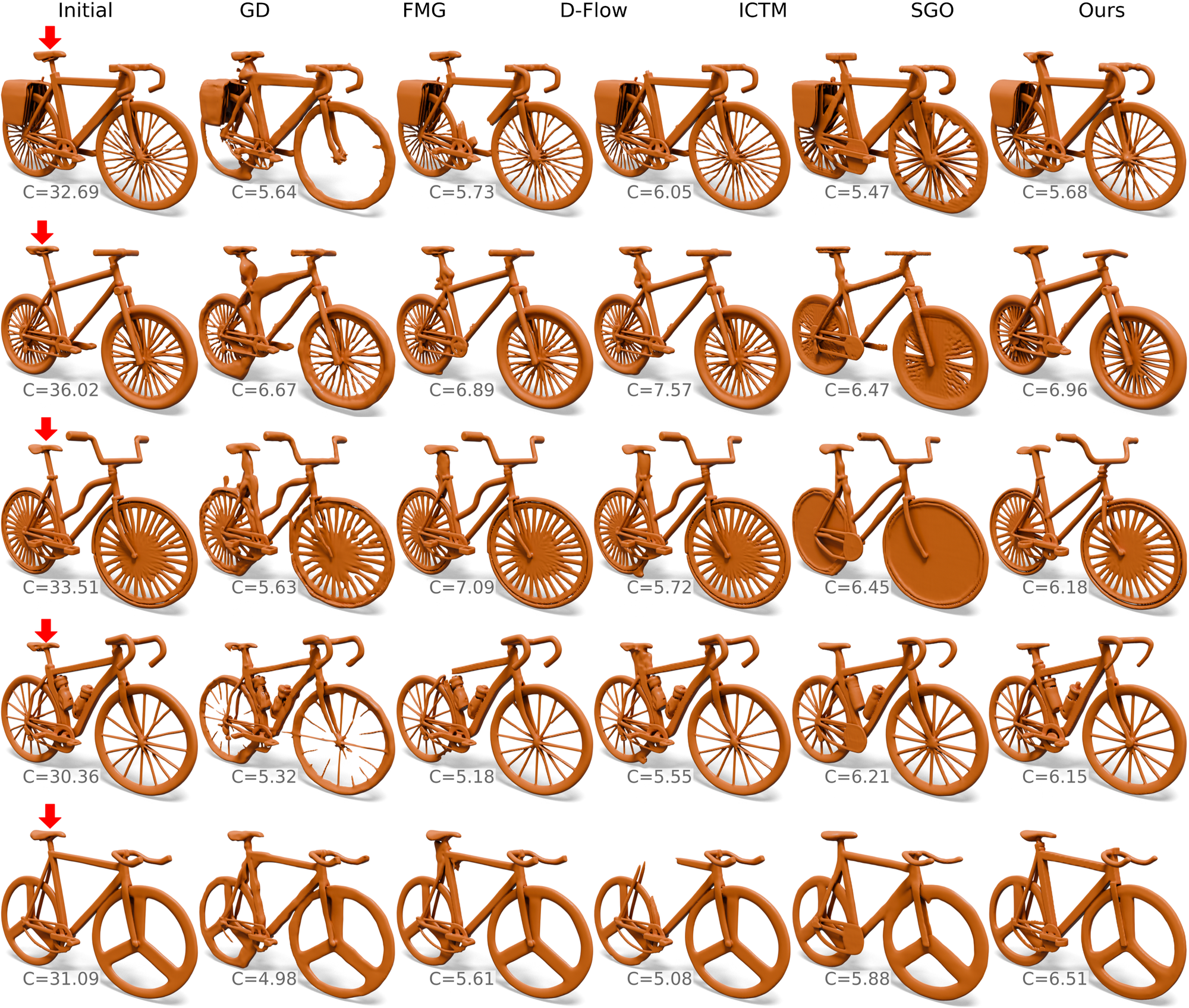} 
\caption{Qualitative result on the BIKED dataset with bicycles.
We show the initial shapes (Init) with the applied force (red arrow) and the optimized shapes produced by different methods (GD, FMG, ICTM, SGO, and Ours).
The compliance ($C$) is reported for each shape, where lower values indicate better structural stability.}
\label{fig:app_ago_bike_shape}
\end{figure}

\newpage
\paragraph{BIKED Dataset.}
In Fig.~\ref{fig:app_ago_bike_shape}, we present visual examples of bicycles optimized on the BIKED dataset.
Similar to the results on the ABO dataset, our method demonstrates superior preservation of the original geometry, adjusting the bicycle frame naturally where needed, while baseline approaches produce blocky, bumpy deformations or missing parts.

\subsection{Additional objective: 2D silhouette matching}
\label{app:silhouette}

The primary goal of our paper is to introduce a robust framework specifically designed for physical 3D shape optimization.
However, to demonstrate that our core idea transfers to other types of objectives that also suffer from manifold drift, we tackle a 2D silhouette matching task on chairs, using a differentiable renderer in a framework conceptually similar to MeshSDF~\cite{Remelli20b}.

\parag{Methodology.}
Starting from an initial latent $z_0$, we decode it into a 3D shape $x$ and extract its binary 2D silhouette $S(x)$ via a differentiable renderer.
We then optimize the latent vector to minimize the $L_2$ distance against a target silhouette $S_{\text{target}}$, derived from a different shape:
\begin{equation}
    \mathcal{L} = \| S(x) - S_{\text{target}} \|_2^2
\end{equation}
We conducted this experiment on the Chair category, using 3DShape2VecSet as the underlying generative model.
Because the objective explicitly matches a geometric target, it acts somewhat like a reconstruction term that structurally grounds the optimization.
In contrast, open-ended objectives, like drag reduction, lack this direct geometric anchor.
To ensure a fair comparison across all methods, we performed this silhouette matching on 100 source-target pairs and computed the Fréchet and Kernel metrics on the resulting 3D shapes, after they reached an average per-pixel error of 0.2\%, a threshold corresponding to convergence where the source and target silhouettes are visually indistinguishable.

\begin{figure}[H]
    \centering
    \includegraphics[width=\linewidth]{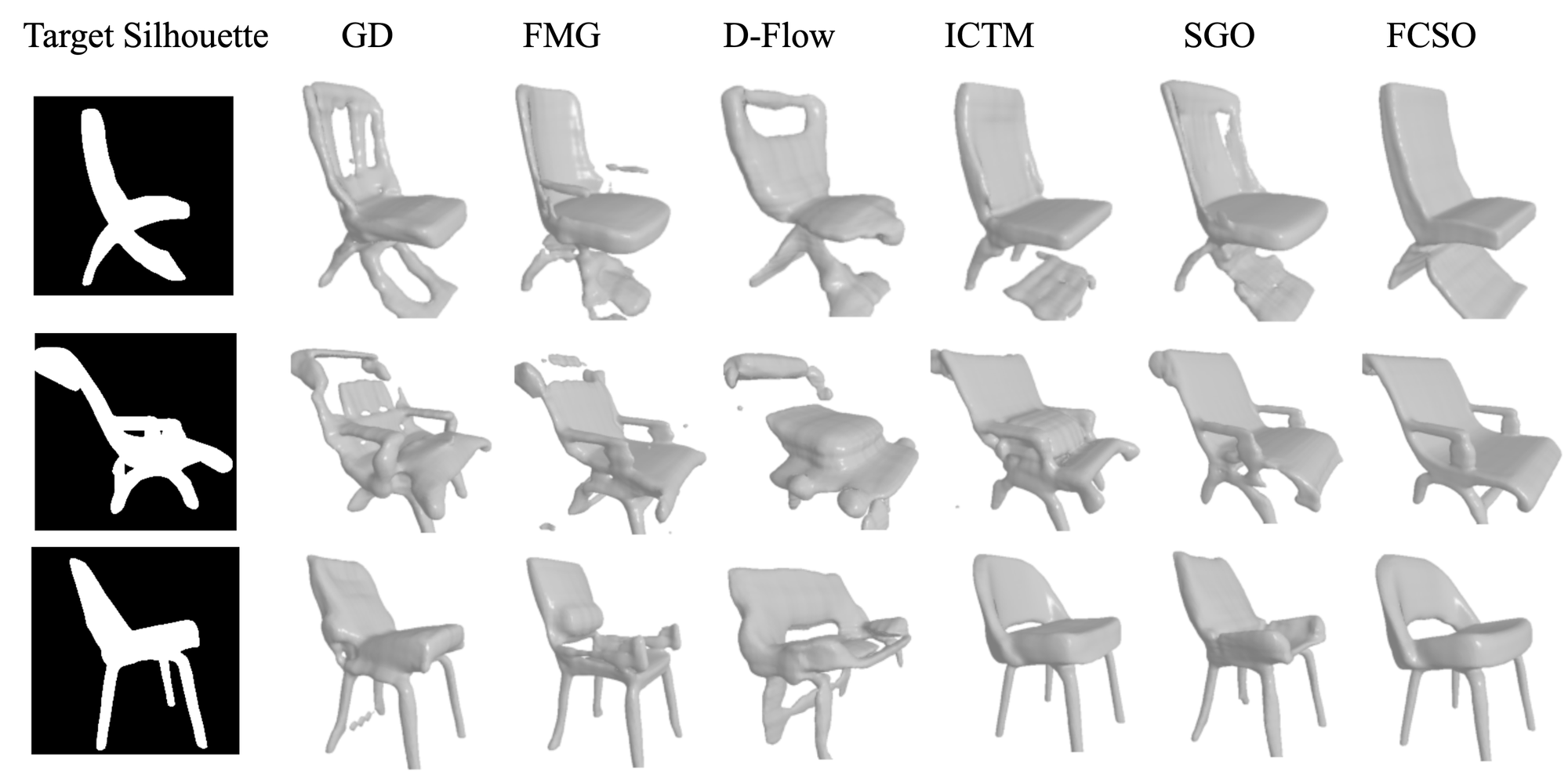}
    \caption{Qualitative results for the 2D silhouette matching experiment. While standard gradient descent often produces adversarial 3D geometries (e.g., floating parts on chairs) to perfectly match the 2D silhouette, our method correctly updates the full 3D structure and stays on the valid shape manifold.}
    \label{fig:silhouette_qualitative}
\end{figure}

\parag{Results.}
As shown in Tab.~\ref{tab:silhouette_chair}, our method consistently outperforms the baselines on the chair dataset.
This category proves highly susceptible to manifold drift, leading to a significant performance gap between FCSO and standard Gradient Descent (GD).
As illustrated in Fig.~\ref{fig:silhouette_qualitative}, because the objective only supervises a 2D projection, baselines tend to converge toward adversarial 3D configurations. While these geometries match the target silhouette from the rendering viewpoint, they fail to form valid 3D shapes.
For instance, the optimizer frequently detaches components, such as armrests, from the main body to force alignment with the target render.
In contrast, our method effectively regularizes the optimization, preserving the structural integrity of the generated shapes.

\begin{table}[htbp]
    \centering
    \caption{Frechet and Kernel metrics for the 2D silhouette matching task on the Chair category (3DShape2VecSet)}
    \label{tab:silhouette_chair}
    \vspace{2mm}
    \small
    \setlength{\tabcolsep}{4pt}
    \begin{tabular}{l c c c c c c}
        \toprule
        \multirow{3}{*}{\textbf{Method}} & \multicolumn{2}{c}{\textbf{PointBert}} & \multicolumn{2}{c}{\textbf{Inception}} & \multicolumn{2}{c}{\textbf{DINOv2}} \\
        & FD $\downarrow$ & KD $\downarrow$ & FD $\downarrow$ & KD $\downarrow$ & FD $\downarrow$ & KD $\downarrow$ \\
        & $(\times 10^3)$ & $(\times 10^1)$ & $(\times 10^4)$ & & & \\
        \midrule
        GD     & $78.33$ & $48.66$ & $403.10$ & $21.39$ & $3.19$ & $5.91$ \\
        FMG    & $80.44$ & $55.52$ & $450.29$ & $28.32$ & $3.26$ & $6.29$ \\
        D-Flow & $102.38$ & $63.22$ & $466.12$ & $30.23$ & $3.56$ & $6.98$ \\
        ICTM   & $42.99$ & $21.81$ & $211.84$ & $9.72$ & $2.32$ & $1.72$ \\
        SGO    & $51.13$ & $25.43$ & $255.21$ & $12.93$ & $2.41$ & $2.19$ \\
        \textbf{Ours} & $\mathbf{39.45}$ & $\mathbf{16.51}$ & $\mathbf{199.06}$ & $\mathbf{8.99}$ & $\mathbf{2.00}$ & $\mathbf{1.55}$ \\
        \bottomrule
    \end{tabular}
\end{table}

\subsection{Limitations: minor artifact in high optimization targets}
\label{app:fail}

\begin{figure}[H]
    \centering
    \includegraphics[width=\linewidth]{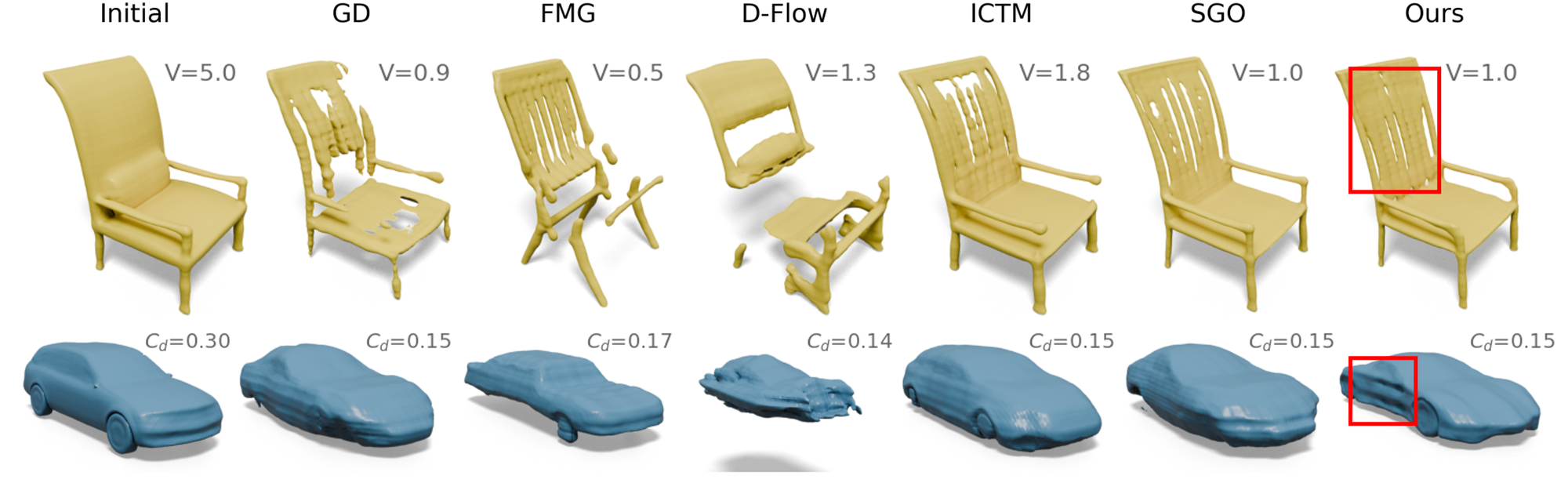}
    \caption{Minor artifacts under high optimization targets across different baselines and our method.}
    \label{fig:app_failure}
\end{figure}

Under high optimization targets, our method can occasionally introduce minor geometric artifacts, as highlighted by the red bounding boxes in Figure~\ref{fig:app_failure}.
However, even in these rare cases, our approach consistently preserves the structure of the shape.
Unlike baselines, a chair remains recognizably a chair, and a car retains its coherent vehicle profile.

\end{document}